\documentclass[reprint,pre]{revtex4-1}
\usepackage{newfloat,algcompatible}
\usepackage{etoolbox}
\usepackage{amsfonts}
\usepackage{amssymb}
\usepackage{graphicx}
\usepackage{mathrsfs}
\usepackage{array}
\usepackage{amsmath}
\usepackage{lscape}
\usepackage{bbold}
\usepackage{ucs}
\usepackage{color}
\usepackage{float}
\usepackage{subcaption}
\usepackage{bm}

\usepackage[utf8]{inputenc}
\usepackage[T1]{fontenc}
\usepackage{caption}
\usepackage{subcaption}

\usepackage{xr}

\makeatletter
\AtBeginDocument{\let\LS@rot\@undefined}
\makeatother

\newcommand{\avdata}[1]{ \left\langle{#1} \right\rangle_{\text{data}} }
\newcommand{\avmodel}[1]{ \left\langle{#1} \right\rangle_{\text{RBM}} }

\newcommand{\beginsupplement}{%
        \setcounter{table}{0}
        \renewcommand{\thetable}{S\arabic{table}}%
        \setcounter{figure}{0}
        \renewcommand{\thefigure}{S\arabic{figure}}%
     }

\begin{document}
%\title{Learning Lattice-Protein Sequence Data with Restricted Boltzmann Machines: \\
%Compositional Regime and Comparative Analysis}
%\title{Compositional analysis with Restricted Boltzmann Machines: Principles and guidelines}
\title{Learning Compositional Representations of Interacting Systems \\with Restricted Boltzmann Machines: Comparative Study of Lattice Proteins}

\author{J\'er\^ome Tubiana, Simona Cocco, R\'emi Monasson}
\affiliation{Laboratory of Physics of the Ecole Normale Sup\'erieure, CNRS \& PSL Research, 24 rue Lhomond, 75005 Paris, France}

\begin{abstract}
A Restricted Boltzmann Machine (RBM) is an unsupervised machine-learning bipartite graphical model that jointly learns a probability distribution over data and extracts their relevant statistical features. As such, RBM were recently proposed for characterizing the patterns of coevolution between amino acids in protein sequences and for designing new sequences. Here, we study how the nature of the features learned by RBM changes with its defining parameters, such as the dimensionality of the representations (size of the hidden layer) and the sparsity of the features. We show that for adequate values of these parameters, RBM operate in a so-called compositional phase in which visible configurations sampled from the RBM are obtained by recombining these features. We then compare the performance of RBM with other standard representation learning algorithms, including Principal or Independent Component Analysis, autoencoders (AE), variational auto-encoders (VAE), and their sparse variants.  We show that RBM, due to the stochastic mapping between data configurations and representations, better capture the underlying interactions in the system and are significantly more robust with respect to sample size than deterministic methods such as PCA or ICA. In addition, this stochastic mapping is not prescribed a priori as in VAE, but learned from data, which allows RBM to show good performance even with shallow architectures. All numerical results are illustrated on  synthetic  lattice-protein data, that share similar statistical features with real protein sequences, and for which ground-truth interactions are known.
\end{abstract}
\maketitle

\section*{Introduction}

Many complex, interacting systems have collective behaviors that cannot be understood based on a top-down approach only. This is either because the underlying microscopic interactions between the constituents of the system are unknown - as in biological neural networks, where the set of synaptic connections are unique to each network - or because the complete description is so complicated that analytical or numerical resolution is intractable - as for proteins, for which physical interactions between amino acids can in principle be characterized, but accurate simulations of protein structures or functions are computationally prohibitive. In the last two decades, the increasing availability of large amounts of data collected by high-throughput experiments such as large scale functional recordings in neuroscience (EEG, Fluorescence imaging,...) \cite{schwarz2014chronic,wolf2015whole}, fast sequencing technologies \cite{finn2015pfam,kolodziejczyk2015technology} (Single RNA seq) or Deep Mutational Scans \cite{fowler2010high} has shed new light on these systems.

Given such high-dimensional data, one fundamental task is to establish a descriptive phenomenology of the system. For instance, given a recording of spontaneous neural activity in a brain region or in the whole brain (e.g. in larval zebrafish), we would like to identify stereotypes of neural activity patterns (e.g. activity bursts, synfire chains, cell-assembly activations, ...) describing the dynamics of the system. This representation is in turn useful to link the behaviour of the animal to its neural state and to understand the network architecture. Similarly, given a Multiple Sequence Alignment (MSA) of protein sequences, i.e. a collection of protein sequences from various genes and organisms that share common evolutionary ancestry, we would like to identify amino acids motifs controlling the protein functionalities and structural features, and identify, in turn, subfamilies of proteins with common functions. One important set of tools for this purpose are unsupervised representation-learning algorithms. For instance, Principal Component Analysis can be used for dimensionality reduction, i.e. for projecting system configurations into a low-dimensional representation, where similarities between states are better highlighted and the system evolution is tractable. Another important example is clustering, which partitions the observed data into different 'prototypes'. Though these two approaches are very popular, they are not always appropriate: some data are intrinsically multidimensional, and cannot be reduced to a low-dimensional or categorical representation. Indeed, configurations can mix multiple, weakly related features, such that using a single global distance metric would be too reductive. For instance, neural activity states are characterized by the clusters of neurons that are activated, which are themselves related to a variety of distinct sensory, motor or cognitive tasks. Similarly, proteins have a variety of biochemical properties such as binding affinity and specificity, thermodynamic stability, or allostery, which are controlled by distinct amino acid motifs within their sequences. In such situations, other approaches such as Independent Component Analysis or Sparse Dictionaries, which aim at representing the data by a (larger) set of independent latent factors appear to be more appropriate \cite{mckeown1998analysis,rivoire2016evolution}.

A second goal is to infer the set of interactions underlying the system's collective behaviour. In the case of neural recordings, we would look for functional connectivity that reflect the structure of the relevant synaptic connections in a given brain state. In the case of proteins, we would like to know what interactions between amino acids shape the protein structure and functions. For instance, a Van Der Waals repulsion between two amino acids is irrelevant if both are far away from one another in the tridimensional structure; the set of relevant interactions is therefore linked to the structure. One popular approach for inferring interactions from observation statistics relies on graphical e.g. Ising or Potts models. It consists in first defining a quadratic log-probability function, then inferring the associated statistical fields (site potentials) and pairwise couplings by matching the first and second order moments of data. This can be done efficiently through various approaches, see \cite{nguyen2017inverse,cocco2018inverse} for recent reviews. The inverse Potts approach, called Direct-Coupling-Analysis \cite{weigt2009identification,morcos2011direct} in the context of protein sequence analysis, helps predict structural contact maps for a protein family from sequence data only. Moreover, such approach defines a probability distribution that can be used for artificial sample generation, and, more broadly, for quantitative modeling, e.g. of mutational fitness landscape prediction in proteins \cite{figliuzzi2016,hopf2017}, or of neural state information content \cite{tkacik2010} or brain states \cite{posani2017functional}. The main drawback of graphical models is that, unlike representation learning algorithms, they do not provide any direct, interpretable insight over the data distribution. Indeed, the relationship between the inferred parameters (fields and couplings) and the typical configurations associated to the probability distribution of the model is mathematically well defined, but intricate in practice. For instance, it is difficult to deduce the existence of data clusters or global collective modes from the knowledge of the fields and couplings. 

An interesting alternative for overcoming this issue are Restricted Boltzmann Machines (RBM). A RBM is a graphical model that can learn both a representation and a distribution of the configuration space, naturally combining both approaches. RBM are bipartite graphical models, constituted by a visible layer carrying the data configurations and a hidden layer, where their representation is formed (Fig.~1(a)); unlike Boltzmann Machines, there are no couplings within the visible layer nor between hidden units. RBM were introduced in \cite{ackley1987learning} and popularized by Hinton et al. \cite{hinton2002training,hinton2006fast} for feature extraction and pretraining of deep networks. More recently, RBM were recently shown to be efficient for modeling coevolution in protein sequences \cite{tubiana2018learning}. The features inferred are sequence motifs related to the the structure, function and phylogeny of the protein, and can be recombined to generate artificial sequences with putative properties. An important question is to understand the nature and statistics of the representation inferred by RBM, and how they depend on the choice of model parameters (nature and number of hidden units, sparse regularization). Indeed, unlike PCA, Independent Component Analysis (ICA) or sparse dictionaries, no constraint on the statistics or the nature of the representations, such as decorrelation, independence, ..., are explicitely enforced in RBM. To answer this question, we apply here RBM on synthetic alignments of Lattice Proteins sequences, for which ground-truth interactions and fitness functions are available. We analyze and interpret the nature of the representations learned by RBM as a function of its defining parameters and in connection with theoretical results on RBM drawn from random statistical ensembles obtained with statistical physics tools \cite{tubiana2017emergence}. Our results are then compared to the other feature extraction methods mentioned above.
%Moreover, unlike in reconstruction-based approaches such as autoencoders or PCA, the hidden unit need not encode all variability of the visible layer but only the correlations between visible units.%; the consequence of which is unknown.

% RBM learn both a representation and a distribution of the configuration space, as i) given a data configuration, the activity in the hidden layer defines a representation of data and ii) marginalizing the joint probability distribution over the hidden layer defines a distribution over the data space. 

The paper is organized as follows. In section I, we formally define RBM and present the main steps of the learning procedure. In Section II, we discuss the different types of representations that can be learned by RBM, with a reminder of recent related results on the different operation regimes, in particular, the so-called compositional phase, theoretically found in Random-RBM ensembles. In Section III, we introduce Lattice Proteins (LP) and present the main results of the applications of RBM to LP sequence data. Section IV is dedicated to the comparison of RBM with other representation learning algorithms, including Principal Component Analysis (PCA), Independent Component Analysis (ICA),  Sparse Principal Component Analysis (sPCA), sparse dictionaries, sparse Autoencoders (sAE), and sparse Variational Autoencoders (sVAE).

%Main content:
%
%Depending on the weight statistics and hidden unit potentials, RBM may produce different representations of the data and different energy landscapes. We now illustrate these different behaviors with RBM trained with various hyperparameters on protein sequence alignments. 
%To this end, we will use synthetic multiple sequence alignements (MSA) of Lattice Proteins (LP). As seen in Fig.~\ref{fig_LP}, MSA of LP share many evolutionary features with MSA of natural proteins. However, unlike natural MSA, LP are well-defined in the sense that data is generated at equilibrium from a ground-truth Hamiltonian. Moreover, the amount of available data is arbitrary, allowing us to factor out from the discussion overfitting problems.

%RBM

\section{Restricted Boltzmann Machines}

\begin{figure*}
\centering
\includegraphics[width = 2\columnwidth]{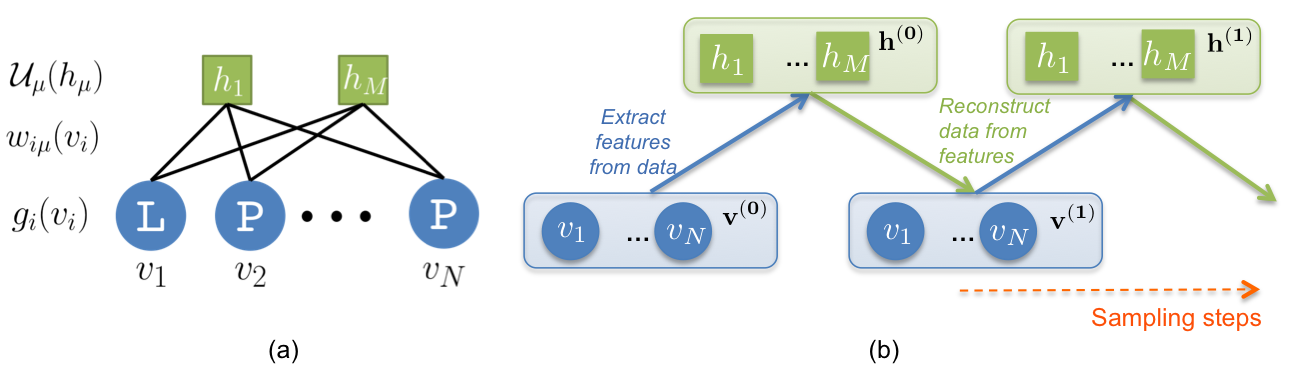}
\caption{{\bf (a)} A restricted Boltzmann machine (RBM) is a two-layer bipartite network, including a visible layer (units $v_i$) carrying the data and a hidden layer (units $h_\mu$) where the data are represented. The dimensions of the data and representation spaces are, respectively, $N$ and $M$. Visible and hidden units are subject to local potentials denoted by, respectively, $g_i$ and ${\cal U}_\mu$; here, we consider that visible units $v_i$ take discrete values (such as amino acids shown in the visible units blue circles) while hidden units $h_\mu$ may take either discrete or real values. {\bf (b)} The alternate Gibbs sampler for sampling from $P({\bf v},{\bf h})$}
\label{fig1}
\end{figure*}

\subsection{Definition}
A Restricted Boltzmann Machine (RBM) is a joint probability distribution over two sets of random variables, the visible layer ${\bf v} = (v_1,..,v_i,...,v_N)$ and hidden layer ${\bf h}= (h_1,..,h_\mu,...,h_M)$. It is formally defined on a bipartite, two-layer graph (Fig.~\ref{fig1} (a)). Depending on the nature of data considered, the visible units $v_i$ can be continuous or categorical, taking $q$ values. Hereafter, we use notations for protein sequence analysis, in which each visible unit represent a site of the Multiple Sequence Alignment and takes $q=21$ values (20 amino acids + 1 alignment gap). The hidden units $h_\mu$ represent latent factors that can be either continuous or binary. Their joint probability distribution is:
\begin{equation}\label{Energy}
P({\bf v},{\bf h}) = \frac 1Z \exp \bigg(\underbrace{ \sum_{i=1}^N g_i(v_i) - \sum_{\mu =1}^M \mathcal{U}_\mu(h_\mu)  + \sum_{i,\mu} h_\mu\, w_{i\mu} (v_i) }_{-E({\bf v},{\bf h})} \bigg)
\end{equation}
where the fields $g_i(v)$ and potentials $\mathcal{U}_\mu$ control the conditional probabilities of the $v_i$, $h_\mu$, and the weights $w_{i\mu}(v)$ couple the visible and hidden variables; the partition function $Z = \sum_{ {\bf v}} \int d{\bf h} e^{-E({\bf v},{\bf h})}$ is defined such that $P$ is normalized to unity. Hidden-unit potential considered here are:
\begin{itemize}
\item The Bernoulli potential: ${\cal U}(0) = 0$, ${\cal U}(1)= u_1$, and ${\cal U}(h) = +\infty$ if $h\ne 0,1$.
\item The quadratic potential: 
\begin{equation} \label{quadpot1}
{\cal U}(h) =  \frac{1}{2} \gamma h^2 + \theta \, h \ ,
\end{equation}
with $h$ real-valued.
\item ReLU potential ${\cal U}(h) =  \frac{1}{2} \gamma h^2 + \theta\, h$, with $h$ real-valued and positive, and ${\cal U}(h)=+\infty$ for negative $h$.
\item The double Rectified Linear Unit (dReLU)  potential:
\begin{equation} \label{doublerelupot}
{\cal U}(h) = \frac{1}{2} \gamma^+ (h^{+})^2 + \frac{1}{2} \gamma^- (h^{-})^2 + \theta^+\, h^+ + \theta^- h^-\ ,
\end{equation}
where $h^+ = \max(h,0)$, $h^- = \min(h,0)$ and $h$ is real-valued.
\end{itemize}

with an associated distribution that can be asymmetric and interpolate between bimodal, gaussian or Laplace-like sparse distributions depending on the choice of the parameters. Bernoulli and quadratic potentials are standard choices in the literature; the so-called double ReLU potential is a more general form, with an associated distribution that can be asymmetric and interpolate between bimodal, gaussian or Laplace-like sparse distributions depending on the choice of the parameters, see Fig.~\ref{potential_transfers}(a).

\subsection{Sampling}
\begin{figure*}
\vspace*{-10mm}
\hspace*{-0.5in}
\includegraphics[scale=1.1]{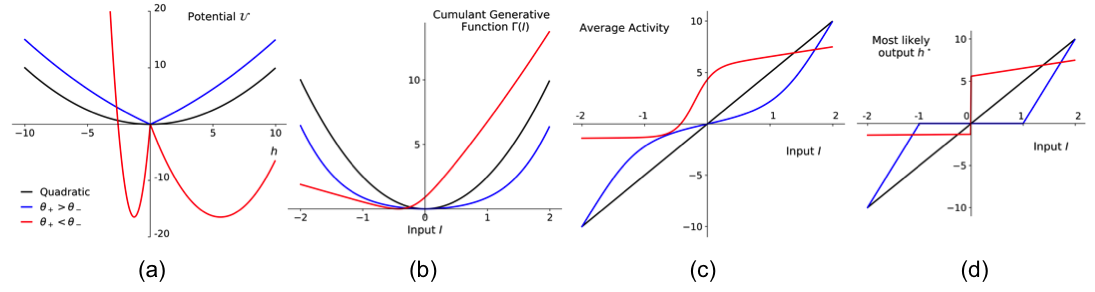}
\caption{ {\bf The double Rectified Linear Units (dReLU) potential} {\bf (a).} Three examples of potentials ${\cal U}$ defining the hidden-unit type in RBM: quadratic potential (black, $\gamma=0.2$), and double Rectified Linear Units (dReLU) potential (blue: $\gamma^+=\gamma^-=0.1$, $\theta^+=-\theta^-=1$; red: $\gamma^+=1$, $\gamma^-=20$, $\theta^+=-6$, $\theta^-=25$). {\bf (b).} Cumulant generative function $\Gamma$ as a  function of the input $I$. {\bf (c).}  Average activity of hidden unit $h$, as a function of the input $I$.  {\bf (d).} Most likely hidden unit value $H(I)$ as a function of the input $I$. }
\label{potential_transfers}
\end{figure*}

Standard sampling from distribution (\ref{Energy}) is achieved by the alternate Gibbs Sampling Monte Carlo algorithm, which exploits the bipartite nature of the interaction graph, see Fig.~\ref{fig1}(b). Given a visible layer configuration $\bf v$, the hidden unit $\mu$ receives the input 
\begin{equation}
I_\mu ({\bf v}) = \sum_i w_{i\mu}(v_i) \ ,
\end{equation}
and the conditional probability of the hidden-unit configuration  is given by $P({\bf h} | {\bf v}) = \prod_\mu P({\bf h_\mu} | {\bf v} )$, where
\begin{equation} \label{cond_proba}
P( h_\mu | {\bf v}) \propto \exp \big(-\mathcal{U}_\mu(h_\mu)  + h_\mu\, I_\mu({\bf v}) \big)\ .
\end{equation}
Similarly, given a hidden layer configuration ${\bf h}$, one can sample from the conditional probability distribution $P({\bf v}|{\bf h})= \prod_i P(v_i |{\bf h} )$, where $P(v_i |{\bf h} )$ is a categorical distribution:
\begin{equation} \label{cond_proba2}
P( v_i | {\bf h}) \propto \exp \big(g_i(v_i) + I_i^v({\bf h}) \big)
\end{equation}
with $I_i^v({\bf h} )= \sum_\mu h_\mu\, w_{i\mu}(v)$. Alternate sampling from $P({\bf h}|{\bf v})$ and $P({\bf v}|{\bf h})$ defines a Markov Chain that eventually converges toward the equilibrium distribution. The potential $\mathcal{U}_\mu$ determines how the average conditional hidden-unit activity $\langle h_\mu \rangle (I)$ and the transfer function $H_\mu(I) = \arg \max P(h_\mu | I)$ vary with the input $I=I_\mu({\bf v})$, see Fig.~\ref{potential_transfers}. For quadratic potential, both are linear functions of the input $I$, whereas, for the ReLU potential, we have $H_\mu(I) = \max \left( \frac{I_\mu - \theta_\mu}{\gamma_\mu}, 0 \right)$. The dReLU potential is the most general form, and can effectively interpolate between quadratic ($\theta_+ = \theta_-$, $\gamma_+ = \gamma_-$), ReLU ($\gamma_- \rightarrow \infty$), and Bernoulli ($\gamma_\pm =\mp \theta_\pm \rightarrow +\infty$) potentials. %For the quadratic potential (Black), the average activity is a linear function of $I$. For $\theta_+>\theta_-$ (Blue), small inputs $I$ barely activate the hidden unit, whereas for $\theta_+<\theta_-$ (Red), the hidden unit essentially binarizes the inputs $I$. 

For $\theta_+ > \theta_-$ (Blue) small inputs $I$ barely activate the hidden unit and the average activity is a soft ReLU, whereas for $\theta_+ < \theta_-$, the average activity rises abruptly at $\frac{\theta_+ \sqrt{\gamma_-} - \theta_-\sqrt{\gamma_+}}{\sqrt{\gamma_+} + \sqrt{\gamma_-}}$ (Red), similarly to a soft thresholding function.

%Finally, we note that given a data configuration ${\bf v}$ on the visible layer, the conditional average hidden activity, $\\langle h_\mu \rangle \big(I_\mu({\bf v})\big)$ effectively defines a representation of the configuration.

\subsection{Marginal distribution}
The marginal distribution over the visible configurations, $P({\bf v})$ is obtained by integration of the joint distribution $P({\bf v},{\bf h})$  over $\bf h$, and reads
\begin{eqnarray} \label{marginal}
P({\bf v}) &=& \int\prod_{\mu=1}^M dh_\mu P({\bf v}, {\bf h}) \\
&=& \frac{1}{Z} \exp \big( \underbrace{\sum_{i=1}^N g_i(v_i) + \sum_{\mu=1}^M \Gamma_\mu\big(I_\mu ({\bf v})\big)}_{-E_\text{eff}({\bf v})} \bigg) \nonumber
\end{eqnarray}
where $\Gamma_\mu(I) = \log \left[ \int dh \, e^{-{\cal U}_\mu(h) + h \,I} \right]$ is the cumulant generative function, or log Laplace transform, associated to the potential ${\cal U}_\mu$ (Fig.~\ref{potential_transfers}(c)). By construction, $\Gamma_\mu'(I) = \langle h_\mu\rangle(I)$ and $\Gamma_\mu''(I)= \langle h_\mu^2\rangle(I) - \langle h_\mu\rangle(I)^2$. 

A special case of interest is obtained for the quadratic potential ${\cal U}(h)$ in (\ref{quadpot1}). The joint distribution $P({\bf v},{\bf h})$ in (\ref{Energy}) is Gaussian in the hidden-unit values $h_\mu$, and the integration can be done straightforwardly to obtain the marginal  distribution over visible configurations, $P({\bf v})$ in (\ref{marginal}). The outcome is
\begin{equation} \label{marginal2}
P({\bf v})  = \frac{1}{Z} \exp \left( \sum_{i} \tilde g_i(v_i) + \frac 12 \sum_{i,j} \tilde J_{ij} (v_i,v_j) \right) \ ,
\end{equation}
with
\begin{eqnarray} 
\tilde g_i(v)  &=& g_i(v)- \frac \theta \gamma \sum _\mu w_{i\mu} (v) \ ,\\
 \tilde J_{ij} (v,v')&=& \frac 1 \gamma \sum _\mu w_{i\mu} (v) w_{j\mu} (v') \ .
 \label{marginal3}
\end{eqnarray}
Hence, the RBM is effectively a Hopfield-Potts model with pairwise interactions between the visible units \cite{barra2012equivalence}. The number of hidden units sets the maximal rank of the interaction matrix $\tilde J$, while the weights vectors attached to the hidden units play the roles of the patterns of the Hopfield-Potts model. In general, for non-quadratic potentials, higher-order interactions between visible units are present. 
We also note that, for quadratic potential, the marginal probability is invariant under rotation of the weights $w_{i\mu}(a) \rightarrow \sum_{\mu'} U_{\mu,\mu'} w_{i\mu'}(a)$; the weights cannot therefore be interpreted separately from each other. In general, this invariance is broken for non-quadratic potential.

\subsection{Learning Algorithm}\label{sec_learning}

Training is achieved by maximizing the average log-likelihood of the data $\avdata{\log P({\bf v})}$, see (\ref{marginal}), %[, where the overbar denotes the average over the data items $\bf v$],
by Stochastic Gradient Descent (SGD). For a generic parameter $\theta$, the gradient of the likelihood is given by: 
\begin{equation} \label{all_Gradients}
\frac{\partial \avdata{\log P({\bf v}) }}{\partial \theta} = -\avdata{\frac{\partial E_{\text{eff}}({\bf v})}{\partial \theta} } + \avmodel{ \frac{\partial E_{\text{eff}}({\bf v})}{\partial \theta} }\ ,
\end{equation}
where $\avdata{{\cal O}}$ and $\avmodel{{\cal O}}$ indicate, respectively,  the averages over the data and model distributions of an observable ${\cal O}$. Evaluating the model averages can be done with Monte Carlo simulations\cite{ackley1987learning,tieleman2008training,desjardins2010tempered,desjardins2010adaptive,salakhutdinov2010learning,cho2010parallel}, or mean-field like approximations \cite{gabrie2015training,tramel2017deterministic}. The gradients of the log-likelihood $\log P({\bf v})$ with respect to the fields $g_i(v)$, couplings $w_{i\mu}(v)$ and the hidden-unit potential parameters that we write generically as ${\bf \xi_\mu}$, therefore read:

\begin{equation} \label{Gradient_g}
\frac{\partial \log P}{\partial g_i(v)} = \avdata{\delta_{v_i,v} } - \avmodel{\delta_{v_i,v}}
\end{equation}
\begin{equation} \label{Gradient_w}
\frac{\partial \log P}{\partial w_{i\mu}(v)} = \avdata{\delta_{v_i,v} \, \Gamma'_\mu \left( I_\mu({\bf v}) \right) } - \avmodel{ \delta_{v_i,v}  \,\Gamma'_\mu \left( I_\mu({\bf v}) \right) }
\end{equation}
\begin{equation} \label{Gradient_xi}
\frac{\partial \log P}{\partial \xi_{\mu}} = \avdata{ \frac{\partial}{\partial {\xi_\mu}}\Gamma_\mu \left( I_\mu({\bf v}) \right) } - \avmodel{ \frac{\partial}{\partial {\xi_\mu}} \Gamma_\mu \left( I_\mu({\bf v}) \right) }
\end{equation}
%\begin{eqnarray} \label{all_Gradients}
%\frac{\partial \log P}{\partial g_i(v)} &=& \avdata{\delta_{v_i,v} } - \avmodel{\delta_{v_i,v}}\ ,\\
%\frac{\partial \log P}{\partial w_{i\mu}(v)} &=& \avdata{\delta_{v_i,v} \, \Gamma'_\mu \left( I_\mu({\bf v}) \right) } - \avmodel{ \delta_{v_i,v}  \,\Gamma'_\mu \left( I_\mu({\bf v}) \right) }\ , \nonumber \\
%\frac{\partial \log P}{\partial \xi_{\mu}} &=& \avdata{ \frac{\partial}{\partial {\xi_\mu}}\Gamma_\mu \left( I_\mu({\bf v}) \right) } - \avmodel{ \frac{\partial}{\partial {\xi_\mu}} \Gamma_\mu \left( I_\mu({\bf v}) \right) } \ . \nonumber
%\end{eqnarray}
Here $\delta_{v_i,v}=1$ if $v_i=v$ and 0 otherwise denotes the Kronecker function. When the likelihood is maximal, the model satisfies moment-matching equations similar to Potts models. In particular, for a quadratic potential the average activity $\Gamma'_\mu$ is linear and (\ref{Gradient_w}) entails that the difference between the second-order moments of the data and model distributions vanishes.

Additionally, there is a formal correspondence between Eqn.~(\ref{Gradient_w}) and other update equations in feature extraction algorithms such as ICA. For instance, in the FastICA \cite{hyvarinen2000independent} formulation, the weight update is given by $\Delta w \propto \avdata{\delta_{v_i,v} \, f\left( I_\mu({\bf v}) \right)}$ where $f$ is the hidden unit transfer function,  followed by an application of the whitening constraint $\avdata{I_\mu({\bf v}) I_\nu({\bf v})} = \delta_{\mu,\nu}$. In both cases, the first step - which is  identical for both methods - drives the weights toward non-Gaussian features (as long as the transfer function $\Gamma'$ resp. $f$ is non-linear), whereas the second one prevents weights from diverging or collapsing onto one another. The gradient of the partition function, which makes the model generative, effectively behaves as a regularization of the weights. One notable difference is that using adaptive dReLU non-linearities allows us to extract simultaneously both sub-Gaussian (with negative kurtosis such as bimodal distribution) and super-Gaussian features (i.e. with positive kurtosis such as sparse Laplace distribution), as well as asymmetric ones. In contrast, in FastICA the choice of non-linearity biases learning toward either sub-Gaussian or super-Gaussian distributions.

%Additionally, there is a formal correspondence between Eqn.~(\ref{Gradient_w}) and other update equations in feature extraction algorithms such as ICA. Indeed, in its Infomax formulation \cite{bell1995information} the updates write: $\avdata{\delta_{v_i,v} \, f\left( I_\mu({\bf v}) \right) } + W^{-1}_{ji}$ 

In addition, RBM can be regularized to avoid overfitting by adding to the log-likelihood penalty terms over the fields and the weights; we use the standard $L_2$ regularization for the fields $\propto \sum_{i,v} g_i(v)^2$, and a $L_1/L_2$ penalty for the weights 
%\begin{equation}\label{regu}
%\Delta \log P =- \frac{\lambda_1^2}{2}\sum_\mu \left( \sum_{i,v} |w_{i\mu}(v)| \right)^2 \ . 
%\end{equation}
\begin{equation}\label{regu}
\mathcal{R} = \frac{\lambda_1^2}{2 N q}\sum_\mu \left( \sum_{i,v} |w_{i\mu}(v)| \right)^2 \ . 
\end{equation}
The latter choice can be explained by writing its gradient $\propto \left( \sum_{i,v} |w_{i\mu}(v)| \right) \text{sign}(w_{i\mu}(v) )$: it is similar to the $L_1$ regularization, with a strength increasing with the weights, hence promoting homogeneity among hidden units. These regularization terms are substracted to the log-likelihood $\log P({\bf v})$ defined above prior to maximization over the RBM parameters. Readers interested in practical details training are referred to the review by Fischer and Igel \cite{fischer2012introduction} for an introduction, and to \cite{tubiana2018learning} for the details of the algorithm used in this paper.

\section{Nature of the representation learnt and weights}

\begin{figure*}
\vspace*{-10mm}
\hspace*{-10mm}
\includegraphics[scale=1.0]{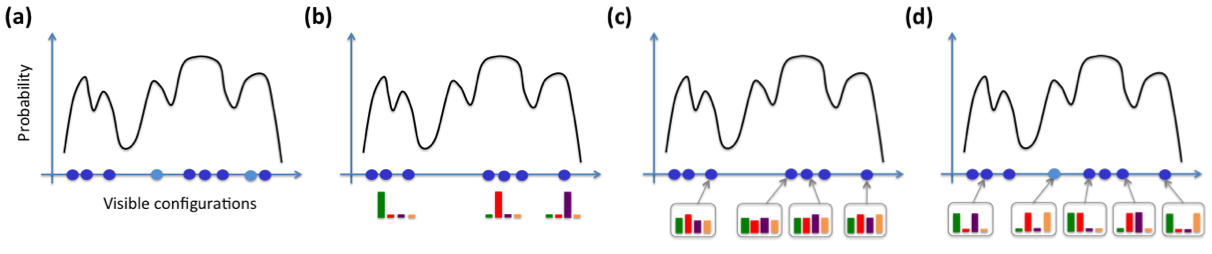}
\caption{{\bf Nature of representations of data with RBM.} {\bf (a).} Data items (dark blue dots) form a very sparse sampling of the high-dimensional distribution (black curve) over visible configurations one wants to reconstruct; many configurations having high probabilities (light blue dots) are not in the data set. {\bf (b).} Prototypical representations: each high-probability portion of the data space activates specifically one hidden unit, while the other units  have very weak activations. Here, hidden units are shown by colored bars, and their activation levels by the heights of the bars. {\bf (c).} Intricate representations: all hidden units have similar amplitudes and play similar roles across data space; visible configurations with high probability are obtained by slightly varying their values. {\bf (d).} Compositional representations: representations are composed of elementary features (attached to hidden units), which can be combinatorially composed to create a large diversity of data items all over the space (light blue dot). The same feature can be active in distinct portions of the configuration space.} 
\label{fig_repr}
\end{figure*}

\subsection{Nature of representations and interpretation of the weights}
Taken together, the hidden units define a probability distribution over the visible layer via (\ref{marginal}) that can be used for artificial sample generation, scoring or Bayesian reconstruction as well as a representation that can be used for supervised learning. However, one fundamental question is how to interpret the hidden units and their associated weights when taken individually. Moreover, what does the model tell us about the data studied?

To this end, we recall first that a data point ${\bf v}$ can be approximately reconstructed from its representation ${\bf h}$ via a so-called linear - non-linear relationship \textit{i.e.} a linear transformation $I_i(v,{\bf h}) = \sum_\mu w_{i\mu}(v) h_\mu$ of the hidden layer activity followed by an element-wise (stochastic) non-linearity $P(v_i | I_i)$, see the conditional distribution $P(v_i|{\bf h})$ of Eqn.~\ref{cond_proba2}. A similar reconstruction relationship also exists in other feature extraction approaches; for instance it is linear and deterministic for PCA/ICA, non-linear deterministic for autoencoders see Section~\ref{sec_comparison}. Owing to this relationship, hidden units may be interpreted as the underlying degrees of freedom controlling the visible configurations. However, such interpretation relies on the constraints imposed on the distribution of $P({\bf h})$, such as decorrelation, independence or sparse activity. In contrast, in RBM the marginal $P({\bf h})$ is obtained by integrating over the visible layer (similarly as $P({\bf v})$) and has no explicit constraints. Therefore, hidden units may be interpreted differently depending on the statistics of $P({\bf h})$. We sketch in Fig.~\ref{fig_repr} several possible scenarios for $P({\bf h})$.

\begin{itemize}
\item{\em Prototypical representations}, see Fig.~\ref{fig_repr}(b). For a given visible layer configuration, about one hidden unit is strongly activated while the others are weakly activated or silent; each hidden unit responds to a localized region of the configuration state. Owing to the reconstruction relation, the weights attached to the hidden unit are therefore ‘prototypes’ data configuration, similar to the centroids obtained with a clustering algorithm.

\item{\em Intricate representations}, see Fig.~\ref{fig_repr}(c). A visible layer configuration activates weakly many (of order $M$) hidden units. Each hidden unit is sensitive to a wide region of the configuration space, and different configurations correspond to slightly different and correlated levels of activation of the hidden units. Taken altogether, the hidden units can be very informative about the data but the individual weights do not have a simple interpretation. \footnote{One extreme example of such intricate representation is the random gaussian projection for compressed sensing \cite{donoho2006compressed}. Provided that the configuration is sparse in some known basis, it can be reconstructed from a small set of linear projections onto random iid gaussian weights $w_{i\mu}$. Although such representation carries all information necessary to reconstruct the signal, it is by construction unrelated to the main statistical features of the data.}

\item{\em Compositional representations}, see Fig.~\ref{fig_repr}(d). A substantial number of hidden units (large compared to one but small compared to the size of the hidden layer) are activated by any visible configuration, while the other units are irrelevant. The same hidden unit can be activated in many distinct parts of the configuration space, and different configurations are obtained through combinatorial choices of the strongly activated hidden units. Conversely, the weights correspond to constitutive ‘parts’ that, once combined, produce a typical configuration. Such compositional representation share similar properties with the ones obtained with sparse dictionaries \cite{olshausen1996emergence,mairal2009online}.
\end{itemize}

Compositional representations offer several advantages with respect to the other types. First, they allow RBM to capture invariances in the underlying distribution from vastly distinct data configurations (contrary to the prototypical regime). Secondly, representations are sparse (as in the prototypical regime), which makes possible to understand the relationship between weights and configurations, see discussion above. Thirdly, activating hidden units of interest (once their weight have been interpreted) allows one to generate configurations with desired properties (Fig.~\ref{fig_repr}(d)). A key question is whether we can force the RBM to generate such compositional representations.

\subsection{Link between model parameters and nature of the representations}
As will be illustrated below, it turns out that all three behaviours can arise in RBM. Here, we argue that what determine the nature of the representation are the global statistical properties of the weights such as their magnitude and sparsity, as well as the number of hidden unit and the nature of their potential. This can be seen from a Taylor expansion of the marginal hidden layer distribution $P(h)$ for the generated data. For simplicity, we focus on the case where $M=2$ and the model is purely symmetrical, with $v_i \in \pm 1$, $g_i(1) = g_i(-1) = 0$, $w_{i\mu}(\pm 1)=\pm w_{i\mu}$, $\gamma_\mu^+ = \gamma_\mu^-=1$, $\theta_\mu^+ = -\theta_\mu^- \equiv \theta_\mu$. For dimensional analysis, we also write $p$ the fraction the weights $w_{i\mu}$ that are significantly non-zero, and $\sim W/\sqrt{N}$ their corresponding amplitude. Then the marginal probability over the hidden layer $P({\bf h})$ is given by:

\begin{equation}
\begin{split}
&\log P({\bf h})=\log \sum_{{\bf v}} P({\bf v},{\bf h}) \equiv L({\bf h}) - \log Z \\ &L({\bf h})= \sum_{\mu=1,2} - \frac{1}{2} h_\mu^2 - \theta_\mu |h_\mu| + \sum_i \log \cosh \left(\sum_{\mu=1,2} w_{i\mu} h_\mu \right) \\
&= \sum_{\mu=1,2} \underbrace{- \frac{1}{2} h_\mu^2}_{SI} - \underbrace{\theta_\mu |h_\mu|}_{SI/SE} + \underbrace{ \frac{1}{2} \left( \sum_i w_{i\mu}^2 \right) h_\mu^2}_{SE} - \underbrace{\frac{1}{12} (\sum_i w_{i\mu}^4) h_\mu^4}_{SI} \\
&+\underbrace{\left(\sum_i w_{i1} w_{i2} \right)  h_1 h_2 }_{CE/CI} - \underbrace{\frac{1}{2} \left( \sum_i w_{i1}^2 w_{i2}^2 \right) h_1^2 h_2^2}_{CI} \\&- \underbrace{\frac{1}{3} \left( \sum_i w_{i1}^3 w_{i2} \right) h_1^3 h_2 - \frac{1}{3} \left( \sum_i w_{i1} w_{i2}^3 \right) h_1^3 h_2}_{CE/CI} 
\end{split}
\end{equation}

%
%\begin{equation}
%\begin{split}
%&\log P({\bf h})=\log \sum_{{\bf v}} P({\bf v},{\bf h}) \equiv L({\bf h}) - \log Z \\ &L({\bf h})= \sum_\mu - \frac{1}{2} h_\mu^2 - \theta_\mu |h_\mu| + \sum_i \log \cosh \left(\sum_\mu w_{i\mu} h_\mu \right) \\
%&= \sum_\mu \underbrace{- \frac{1}{2} h_\mu^2}_{SI} - \underbrace{\theta_\mu |h_\mu|}_{SI/SE} + \underbrace{ \frac{1}{2} \left( \sum_i w_{i\mu}^2 \right) h_\mu^2}_{SE} - \underbrace{\frac{1}{12} (\sum_i w_{i\mu}^4) h_\mu^4}_{SI} \\
%&+ \sum_{\mu \neq \mu’} \underbrace{\frac{1}{2}  \left(\sum_i w_{i\mu} w_{i\mu’} \right)  h_\mu h_{\mu’} }_{CE/CI} - \underbrace{\frac{1}{4} \left( \sum_i w_{i\mu}^2 w_{i\mu’}^2 \right) h_\mu^2 h_{\mu’}^2}_{CI} \\&- \underbrace{\frac{1}{3} \left( \sum_i w_{i\mu}^3 w_{i\mu'} \right) h_\mu^3 h_{\mu'}}_{CE/CI} - \frac{1}{12} \sum_{\mu \neq \mu' \neq \mu''} \left( \sum_i w_{i\mu}^2 w_{i\mu'} w_{i\mu''} \right) h_\mu^2 h_{\mu'} h_{\mu''} \\&- \frac{1}{12} \sum_{\mu \neq \mu' \neq \mu'' \neq \mu'''} \left( \sum_i w_{i\mu} w_{i\mu'} w_{i\mu''} w_{i\mu'''} \right) h_\mu h_{\mu'} h_{\mu''} h_{\mu'''} + ...
%\end{split}
%\end{equation}
where we have discarded higher order terms. This expansion shows how hidden units effectively interact with one another, via self-excitation  (SE), self-inhibition (SI), cross-excitation (CE) or cross inhibition (CI) terms. Some key insights are that:

\begin{itemize}
\item Large $h$ values arise via self-excitation, with a coefficient in $\sum_i w_{i,\mu}^2 \sim p W^2$.
\item The dReLU threshold acts either as a self-inhibition that can suppress small activities ($\theta_\mu>0$) or as self-excitation that enhances them.

\item Hidden unit interactions (hence correlations), arise via their overlaps $\sum_{i} w_{i\mu} w_{i\mu’} \sim p^2 W^2$.
\item The non-gaussian nature of the visible units induces an effective inhibition term between each pair of hidden units, whose magnitude $\sum_i w_{i\mu}^2 w_{i\mu'}^2 \sim \frac{p^2 W^4}{N}$ essentially depends on the overlap between the supports of the hidden units. We deduce that i) the larger the hidden layer, the stronger this high-order inhibition, and that ii) RBM with sparse weights (order $p$ non-zero coefficients) have significantly weaker overlaps (of order $p^2$), and therefore have much fewer cross-inhibition.
\end{itemize}

Depending on the parameter values, several different behaviours can therefore be observed. Intuitively, the i) prototype, ii) intricate and iii) compositional representation correspond to the cases where i) strong self excitation and strong cross-inhibition result in a winner-take-all situation where one hidden unit is strongly activated and ‘inhibits’ the others. ii) Cross-inhibition dominates, and all hidden units are weakly activated. iii) Strong self-excitation and weak cross-inhibition (e.g. due to sparse weights) allows for several hidden units to be simultaneously strongly activated

Though informative, the above expansion is only valid for small $W$/small $h$, whereas hidden units may take large values in practice; a more elaborate mathematical analysis is therefore required and is presented now.

\subsection{Statistical mechanics of random-weights RBM}
The different representational regimes shown in Fig.~\ref{fig_repr} can be observed and characterized analytically in a simple statistical ensembles of RBM controlled by a few structural parameters  \cite{tubiana2017emergence}:

\begin{itemize}
\item the aspect ratio $\alpha= \frac{M}{N}$ of the RBM;
\item the threshold $\theta_\mu=\theta$ of the ReLU potential acting on hidden units;
\item the local field $g_i=g$ acting on visible units;
\item the sparsity $p$ of the weights $w_{i\mu}$ independently and uniformly drawn at random from the distribution \cite{agliari2012multitask}
\begin{equation}\label{defwp}
w_{i\mu} = \left\{ \begin{array} { c c c}
0 & \text{with probability} & 1 - \frac p2 \ ,\\
\frac W{\sqrt N} & \text{with probability} & \frac p2 \ ,\\
-\frac W{\sqrt N} & \text{with probability} & \frac p2 \ .
\end{array} \right.
\end{equation}
\end{itemize}
Depending on the values of these parameters the RBM may operate, at low temperature {\em i.e.} for $W^2 p \gg 1$, in one of the three following phases when the size $N$ becomes very large:
\begin{itemize}
\item{\em Ferromagnetic phase:} One hidden unit, say, $\mu=1$, receives a strong input 
\begin{equation}
I_1 = \sum _{i} w_{i1} \langle v_i\rangle
\end{equation}
of the order of $\sqrt N$; the other hidden units receive small (of the order of $\pm 1$) inputs and can be silenced by an appropriate finite value of the threshold $\theta$. Hidden unit 1 is in the linear regime of its ReLU activation curve, and has therefore value $h_1 = I_1$. In turn, each visible unit receives an input of the order of $h_1\times w_{i1}$ from the strongly activated hidden unit. Due to the non-linearity in the activitation curve of ReLU and the presence of the threshold $\theta$, most of the other hidden units are silent, and do not send noisy inputs to the visible units. Hence visible unit configurations $\{v_i\}$ are strongly magnetized along the pattern $\{w_{i1}\}$. This phase is the same as the one found in the Hopfield model at sufficiently small loads $\alpha$ \cite{amit1985storing}. 
\item{\em Spin-glass phase:} If the aspect ratio $\alpha$ of the machine is too large, and the threshold $\theta$ and the visible field $g$ are too small, the RBM enters the spin-glass phase. All inputs $I_\mu$ to the visible units are of the order of $\pm 1$; visible units are also subject to random inputs $I_i$ of the order of unity, and are hence magnetized as expected in a spin glass.  
\item{\em Compositional phase:} At large enough sparsity, {\em i.e.} for small enough values of $p$, see (\ref{defwp})), a number $L\sim \frac \ell p$ of hidden units have strong magnetizations of the order of $m\sim p\times\sqrt N$, the other units being shutdown by choices of the threshold $\theta \sim \sqrt p$. Notice that $L$ is large compared to 1, but small compared to the hidden-layer size, $M$; the value of $\ell$ is determined through minimization of the free energy \cite{tubiana2017emergence}. The input $I_i$ onto visible units is of the order of $m\times w\times L \sim p$. The mean activity in the visible layer is fixed by the choice of the visible field $g\sim p$. This phase requires low temperatures, that is weight squared amplitudes $W^2\sim \frac 1p$ at least.
Mathematically speaking, these scalings are obtained when  the limit $p\to 0$  is taken after the 'thermodynamic' limit $N,M\to\infty$ (at fixed ratio $\alpha=M/N)$.
\end{itemize}
Although RBM learnt from data do not follow the same simplistic weight statistics and deviations are expected, the general idea that each RBM learn a decomposition of samples into building blocks was observed on MNIST, a celebrated data set of handwritten digits, and is presented hereafter on in silico protein families.

\subsection{Summary}
To summarize, RBM with non-quadratic potential and sparse weights, as enforced by regularization learn compositional representation of data. In other words, enforcing sparsity in weights results in enforcing  sparsity in activity, in a similar fashion to sparse dictionaries. The main advantages of RBM with respect to the latter are that unlike sparse dictionaries, RBM defines a probability distribution - therefore allowing sampling, scoring, Bayesian inference,... - and that the representation is a simple linear - non-linear transformation, instead of the outcome of an optimization.

Importantly, we note that the R-RBM ensemble analysis only shows that sparsity is a sufficient condition for building an energy landscape with a diversity of gradually related attractors; such landscape could also be achieved in other parameter regions e.g. when the weights are correlated. Therefore, there is no guarantee that training a RBM without regularization on a compositional data set (e.g. generated by a R-RBM in the compositional phase) will result in sparse weights and a compositional representation.

Since we can enforce via regularization a compositional representation, what have we learnt about the data itself? The answer is that enforcing sparsity does not come at the same cost in likelihood for all data sets. In some cases, e.g. for data constituted by samples clustered around prototypes, weight sparsity yields a significant misfit with respect to the data. In other cases such as Lattice Proteins presented below, enforcing sparsity comes at a very weak cost, suggesting that these data are intrinsically compositional. More generally, RBM trained with fixed regularization strength on complex data sets may exhibit heterogeneous behaviors, with hidden units behaving as compositional and other as prototypes, see some examples in \cite{tubiana2018learning}.

%LP
%
\section{Results}
\begin{figure*}
\begin{center}
\includegraphics[scale=1.0]{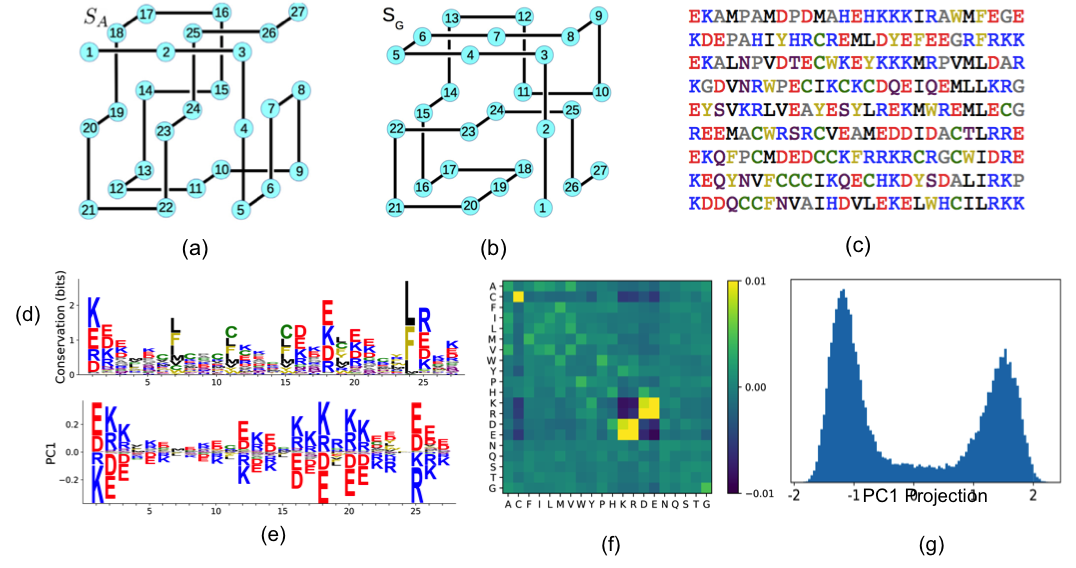}
\end{center}
\caption{ {\bf Main features of Lattice Protein (LP) sequence alignments.} Two examples of structures, $S_A$ and $S_G$, are shown in, respectively, panels {\bf (a)} and {\bf (b)}. {\bf (c).} A subset of the $36,000$ sequences generated by Monte Carlo that fold specifically in $S_A$ ($p_{nat}>0.995$), \textit{i.e.} with much lower energy when folded in $S_A$ {\bf (a)} than in other structures, such as $s_G$ {\bf (b)}. Color code for amino acids: red = negative charge (E,D), blue = positive charge  (H, K, R), purple =  non-charged polar (hydrophilic) (N, T, S, Q), yellow = aromatic (F, W, Y), black = aliphatic hydrophobic  (I, L, M, V), green = cysteine (C), grey = other, small (A, G, P). {\bf (d).} Position weight matrix of the sequence alignment. For each site $i$ with observed amino acid frequency $f_i(v)$, the total height denotes the conservation score, $\log 20 +\sum_v f_i(v) \log f_i(v)$, and the height of each letter is proportional to $f_i(a)$. LP alignment feature non-conserved and partly conserved sites. {\bf (e).} Weight logo of the first principal component, PC1; The heights of letters in panel {\bf (e)} are proportional to the corresponding principal component coefficients, $w_{i}^{\text{PC1}}(v)$. Positive and negative coefficients are respectively above and below the 0 axis {\bf (f).} Average covariance between pairs of sites in contact, $C(v,v') = \frac{1}{28} \sum_{i,j} c_{ij}^{(S)} \, (f_{ij}(v,v')-f_i(v)f_j(v'))$, reflecting the main physical properties through which amino acids interact. 
{\bf (g)} Histogram of the projections of sequences along PC1.  As for natural proteins, LP alignments cluster into subfamilies.}
 \label{fig_LP}
\end{figure*}

\subsection{Lattice proteins: model and data}

Lattice-protein (LP) models were introduced in the $'90$s to investigate the properties of proteins \cite{shakhnovich1990enumeration}, in particular how their structure depend on their sequences. They were recently used to benchmark graphical models inferred from sequence data \cite{jacquin2016benchmarking}. In the version considered here, LP include 27 amino acids and fold on a $3 \times 3\times 3$ lattice cube  \cite{shakhnovich1990enumeration}. There are  ${\cal N}=103,406$ possible folds (up to global symmetries), \textit{i.e.} self-avoiding conformations of the 27 amino acid-long chains on the cube.  

Each sequence ${\bf v}=(v_1,v_2,...,v_{27})$  is assigned the energy
\begin{equation}
{\cal E}({\bf v};S) = \sum_{i<j} c_{ij}^{(S)}\; \epsilon(v_i,v_j)\ .
\label{energy}
\end{equation}
\noindent 
when it is folded in structure $S$. In the formula above, $c^{(S)}$ is the contact map of $S$. It is a $27\times27$ matrix, whose entries are $c^{(S)}_{ij}=1$ if the pair of sites ${ij}$ is in contact in $S$, \textit{i.e.} if $i$ and $j$ are nearest neighbors on the lattice, and  $c^{(S)}_{ij}=0$  otherwise. The pairwise energy $\epsilon(v_i,v_j)$ represents the amino acid physico-chemical interaction between the amino acids $v_i$ and $v_j$ when they are in contact; its value is given by the Miyazawa-Jernigan (MJ) knowledge-based potential \cite{miyazawa1996residue}. 

The probability that the protein sequence $\bf v$ folds in one of the structures, say, $S$, is 
 \begin{equation}
 P_{nat}({\bf v} ; S)=\frac{\displaystyle e^{-{\cal E}({\bf v} ; S)}}{\displaystyle \sum_{S'=1}^{\cal N} e^{-{\cal E}({\bf v}; S')}} \ ,
 \label{pnat}
 \end{equation}
where the sum at the denominator runs over all ${\cal N}$ possible structures, including the native fold $S$. We consider that the sequence $\bf v$ folds in $S$ if $P_{nat}({\bf v} ; S)>0.995$.

A collection of 36,000 sequences $\bf v$ that specifically fold in structure $S_A$, {\em i.e.} such that $P_{nat}({\bf v};S_A)>0.995$, were generated by Monte Carlo simulations as described in \cite{jacquin2016benchmarking}. As real MSA, Lattice-Protein data feature short- and long-range correlations between amino acid on different sites, as well as high-order correlations that arise from competition between folds \cite{jacquin2016benchmarking}, see Fig.~\ref{fig_LP}.

\subsection{Representations of lattice proteins by RBM: interpretation of weights and generative power}

\begin{figure}
\hspace{-0in}
\centering
\includegraphics[width=1\columnwidth]{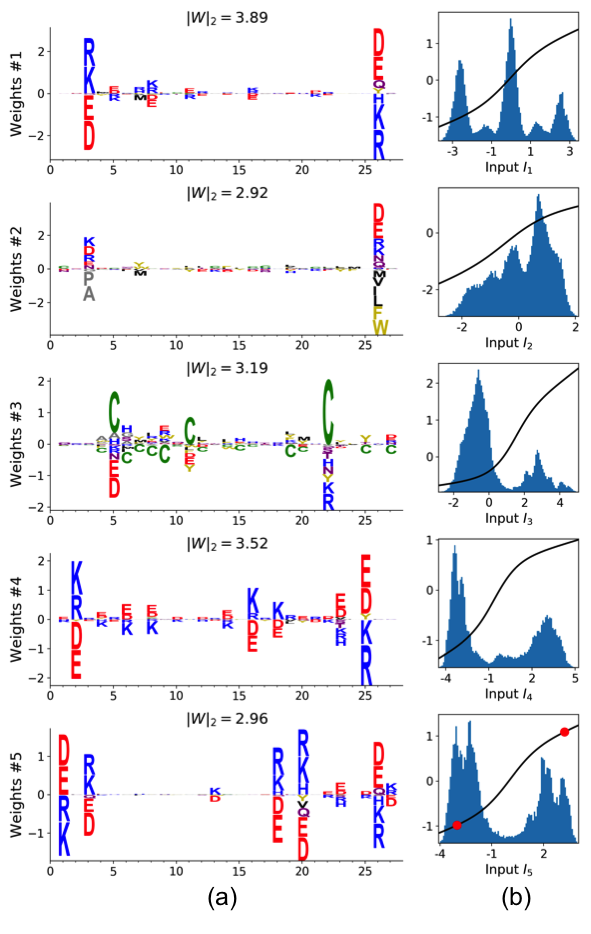}
\caption{{\bf Modeling Lattice Proteins with RBM} {\bf A.} Five weight logos, visualizing the weights $w_{i\mu}(v)$ attached to five selected hidden units. {\bf B} Distribution of inputs received by the corresponding hidden units, and conditional mean activity (full line and left scale). }
\label{RBM_Features_LP}
\end{figure}

We now learn RBM from the LP sequence data, with the training procedure presented in Section \ref{sec_learning}. The visible layer includes $N=27$ sites, each carrying Potts variables $v_i$ taking 20 possible values. Here, we show results with $M=100$ dReLU hidden units, with a regularization $\lambda_1^2=0.025$, trained on a fairly large alignment of $B=36,000$ sequences. We present in Fig.~\ref{RBM_Features_LP} a selection of structural LP features inferred by the model, see \cite{tubiana2018learning} for more features. For each hidden unit $\mu$, we show in panel A the weight logo of $w_{i\mu}(v)$ and in panel B the distribution of its hidden unit input $I_\mu$, as well as the conditional mean activity $\langle h_\mu\rangle(I_\mu)$. Weights have significant values on a limited number of sites only, which makes their interpretation easier. 

As seen from Fig.~\ref{RBM_Features_LP}(a), weight 1 focuses mostly on sites 3 and 26, which are in contact in the structure (Fig.~\ref{fig_LP}(a)) and are not very conserved (sequence logo in Fig.~\ref{fig_LP}(d)). Positively charged residue (H,R,K) have a large positive (resp. negative) component on site 3 (resp. 26), and negatively charged residues (E,D) have a large negative (resp. positive) components on the same sites. The histogram of the input distribution in Fig.~\ref{RBM_Features_LP}(b) shows three main peaks in the data. Since $I_1(v) = \sum_i w_{i1}(v_i)$, the peaks (i) $I_1 \sim 3$, (ii) $I_1 \sim -3$, and (iii) $I_1 \sim 0$ correspond to sequences having, respectively, (i) positively charged amino acids on site 3 and negatively charged amino acids on site 26 (ii) conversely, negatively charged amino acids on site 3 and positively charged on site 26, and (iii) identical charges or non-charged amino acids. Weight 2 also focuses on sites 3 and 26. Its positive and negative components correspond respectively to charged (D,E,K,R,H) or hydrophobic amino acids (I,L,V,A,P,W,F). The bulk in the input distribution around $I_2 \sim -2$ therefore identifies sequences having hydrophobic amino acids at both sites, whereas the positive peak corresponds to electrostatic contacts as the ones shown in weight 1.
The presence of hidden units $1$ and $2$ signals an \textit{excess} of sequences having significantly high $|I_1|$ or $|I_2|$ compared to an independent-site model. Indeed, the contribution of hidden unit $\mu=1,2$ to the log-probability is $\Gamma (I_\mu) \sim I_\mu^2$, since the conditional mean $\langle h_\mu\rangle(I_\mu)=\Gamma'(I_\mu)$ is roughly linear (Fig.~\ref{RBM_Features_LP}(b)). In other words, sequences folding into $S_A$ are more likely to have complementary residues on sites 3 and 26 than would be predicted from an independent-site model (with the same site-frequencies of amino acids). 

Interestingly, RBM can extract features involving more than two sites. Weight 3 is located mainly on sites 5 and 22, with weaker weights on sites 6, 9, 11. It codes for a cysteine-cysteine disulfide bridge located on the bottom of the structure and present in about a third of the sequences ($I_5 \sim 3$). The weak components and small peaks $I_5 \sim 4$ also highlight sequences with a 'triangle' of cysteines 5-11-22 (see structure A). We note however that this is an artifact of the MJ-based interactions in Lattice Proteins, see (\ref{energy}), as a real cysteine amino acid may form only one disulfide bridge.

Weight 4 is an extended electrostatic mode. It has strong components on sites 23,2,25,16,18 corresponding to the upper side of the protein (Fig.~\ref{fig_LP}(a)). Again, these five sites are contiguous on the structure, and the weight logo indicates a pattern of alternating charges present in many sequences ($I_4\gg 0$ and $I_4\ll 0$).

The collective modes defined by RBM may not be contiguous on the native fold. Weight 5 codes for an electrostatic triangle 20-1-18, and the electrostatic 3-26, which is far away from the former. This indicates that despite being far away, sites 1 and 26 often have the same charge. The latter constraint is not due to the native structure $S_A$ but impedes folding in the `competing' structure, $S_G$, in which sites 1 and 26 are neighbours (Fig.~\ref{fig_LP}(b)). Such negative design was also reported through analysis with pairwise model \cite{jacquin2016benchmarking}.

\subsection{Lattice Protein sequence design}

\begin{figure}
\hspace{-0.3in}
\centering
\includegraphics[width=1.1\columnwidth]{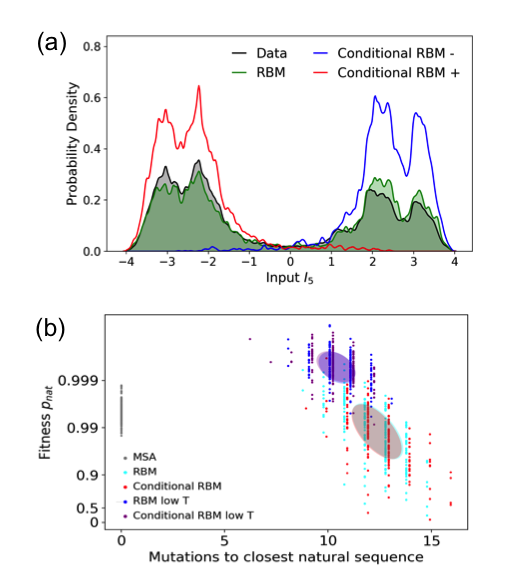}

\caption{{\bf Sequence generation with RBM} {\bf (a).} Distribution of hidden unit input $I_5$, for original sequences (black), RBM-generated sequences (green) and conditional RBM-generated sequences (red,blue). The latter are obtained by sampling while fixing the activity of $h_5$ to one of the two values shown in red in Fig.~\ref{RBM_Features_LP}(b). {\bf (b)} Scatter plot of the number of mutations to the closest natural sequence vs probability to fold into $S_A$, for natural (gray) and artificial (colored) LP sequences. Sequences are generated using regular RBM sampling, conditional RBM sampling and/or low temperature sampling (see Methods of \cite{tubiana2018learning}). The non-linear scale $y = -\log(1-P_{nat})$ was used; The ellipsis indicates best fitting gaussian.}
\label{RBM_Generation_LP}
\end{figure}

The interpretability of the hidden units allows us to design new sequences with the right fold. We show in Fig.~\ref{RBM_Generation_LP}A an example of conditional sampling, where the activity of one hidden unit (here, $h_5$) is fixed. This allows us to design sequences having either positive or negative $I_5$, depending on the sign of $h_5$, hence a subset of five residues with opposite charges. In this example, biasing can be useful e.g. to find sequences with prescribed charge distribution. More broadly, it was reported in \cite{tubiana2018learning} that hidden units can have a structural or functional role, e.g. in term of loop size, binding specificity, allosteric interaction... Conditional sampling allows one in principle to tune these properties at will. As seen from Fig.~\ref{RBM_Generation_LP}(b), the sequences generated have both (i) low sequence similarity to the sequences used in training, with about 40\% sequence difference to the \textit{closest} sequence in the data, and (ii) high probability $P_{nat}$ (\ref{pnat}) to fold into $S_A$. Interestingly, low temperature sampling , e.g. sampling from  $\propto P({\bf v})^2$, can produce sequences with higher $P_{nat}$ than all the sequences used for training. In real protein design setups, this could correspond to higher binding affinity, or better compromises between different target functions.

\subsection{Representations of lattice proteins by RBM: effect of sparse regularization}
\begin{figure*}
\centering
\includegraphics[scale=1.2]{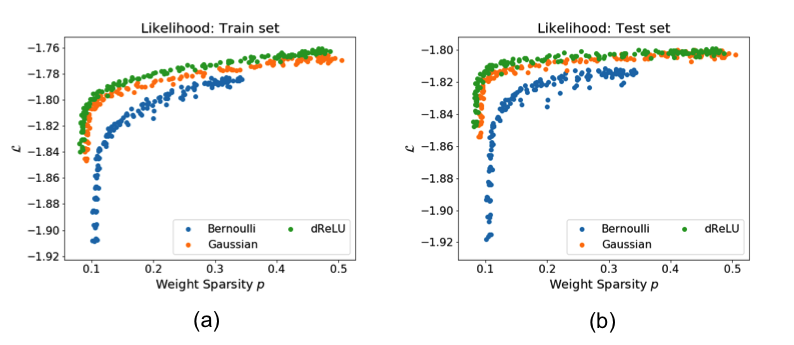}
\caption{Sparsity-performance trade-off on Lattice Proteins. Scatter plots of Log-likelihood vs. Weight sparsity for the training {\bf (a)} and test {\bf (b)} sets. Each point corresponds to a RBM trained with a different regularization parameter.}
\label{sparsity_performance}
\end{figure*}

We now repeat the previous experiment with varying regularization strength, as well as for various potentials.
%The results shown above were obtained for a specific value of the sparse regularization, but how do they vary with the regularization strength? We repeat the training for a range of regularization strengths, as well as for various potentials. 
To see the effect of the regularization on the weights, we compute a proxy $p$ for the weight sparsity, see definition in Appendix. We also evaluate the likelihood $\mathcal{L}$ of the model on the train and a held-out test set; to do so, we estimate the partition function via the Annealed Importance Sampling algorithm \cite{neal2001annealed,salakhutdinov2008quantitative}. We show in Fig.~\ref{sparsity_performance} the Sparsity-Likelihood plot. 

Without sparse regularization, the weights are not sparse and the representation is intricate. As expected, increasing the regularization strength results in fewer non-zero weights and lower likelihood on the training set. Somewhat surprisingly, the test set likelihood decays only mildly for a while, even though the sample is very large such that no overfitting is expected. This suggests that there is a large degeneracy of maximum likelihood solutions that perform equally well on test data; sparse regularization allows us to select one for which the representation is compositional and the weights can be easily related to the underlying interactions of the model.  
Beyond some point, likelihood decreases abruptly because hidden units cannot encode all key interactions anymore. Choosing a regularization strength such that the model lies at the elbow of the curve, as we did in previous section, is a principled way to obtain a model that is both accurate and interpretable.

\subsection{Representations of lattice proteins by RBM: effects of the size of the hidden layer}
\begin{figure*}
\hspace*{-0.5in}
\centering
\includegraphics[scale=0.4]{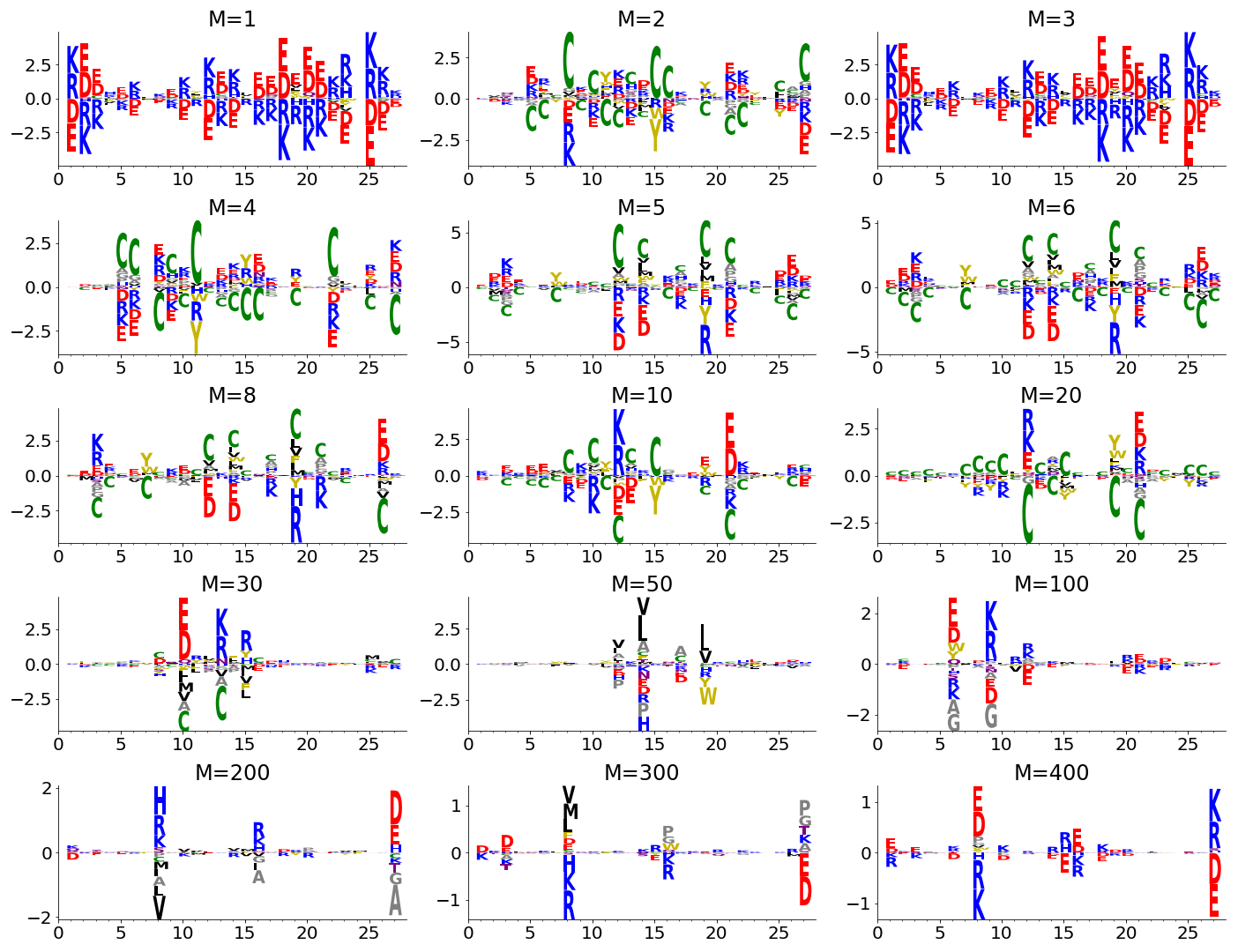}
\caption{Typical features learnt by RBMs on Lattice Proteins for $M\in [1,400]$ with fixed regularization strength $\lambda_1^2=0.025$. For each RBM, one 'typical' feature is shown, with sparsity $p_\mu$ closest to the median sparsity $p$, see definition of $p_\mu$ in Eqn.~\ref{sparsity} of Annex}
 \label{vary_nh}
\end{figure*}

We now study how the value of $M$ impacts the representations of the protein sequences. We repeat the training with dReLU RBM fixed regularization $\lambda_1^2 = 0.025$ for $M$ varying between 1 and 400, and show in Fig.~\ref{vary_nh} one typical weight ${\bf w}_\mu$ learnt by dReLU RBMs for $M$ varying between 1 and 400. We observe different behaviours, as the value of $M$ increases:

For very low $M$, the weight vectors are extended over the whole visible layer. For $M=1$, the unique weight vector capture electrostatic features -- its entries are strong on charged amino acids, such as K, R (positive charges) and E, D (negative charges)--, and is very similar to the top component of the correlation matrix of the data, compare top panel in Fig.~\ref{vary_nh} and Fig.~\ref{fig_LP}(f). Additional hidden units, see panels on the second line of  Fig.~\ref{vary_nh} corresponding to $M=2$, capture other collective modes, here patterns of correlated Cystein-Cystein bridges across the protein. Hence, RBM can be seen as a non-linear implementation of Principal Component Analysis. A thorough comparison with PCA will be presented in Section \ref{sec_comparison}.

\begin{figure*}
\hspace*{-0.8in}
\includegraphics[scale=1.2]{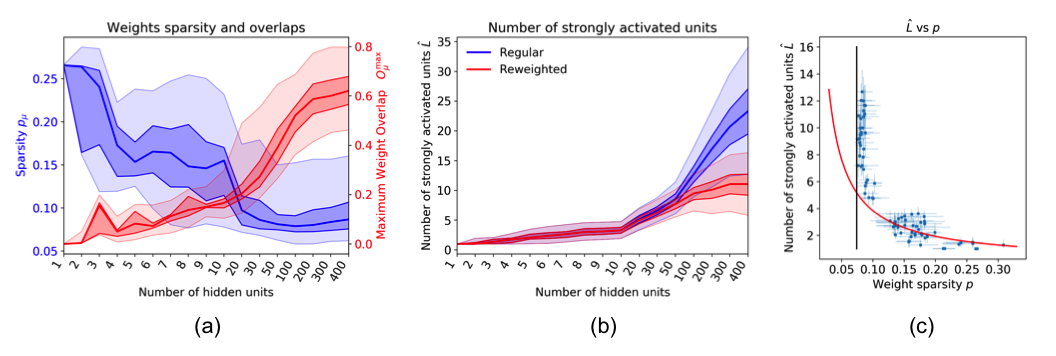}
\caption{(a) Evolution of the distribution of weights sparsity $p_\mu$ and maximum overlaps $O^{\max}_\mu$ as functions of number $M$ of hidden units. (b) Evolution of the number of strongly activated hidden units as function of the number $M$ of hidden units. Solid lines indicate median, and colored area indicate 33$\%$,66$\%$ (dark) and 10$\%$, 90$\%$ quantiles (light).
(c) Scatter plot of the number of strongly activated hidden units against weight sparsity. The vertical bar locates the minimal sparsity value p=2/27, see text.}
\label{compositional_figures}
\end{figure*}

As $M$ increases, the resolution of representations gets finer and finer: units focus on smaller and smaller portions of the visible layer, see Fig.~\ref{vary_nh}. Nonzero entries are restricted to few sites, often in contact on the protein structure, see Fig.~\ref{fig_LP}, fold $S_A$. We introduce a proxy $p$ for the sparsity of the weight based on inverse participation ratios of the entries $w_{i\mu}(v)$, see Appendix. The behavior of $p$ as a function of $M$ is shown in Fig.~\ref{compositional_figures}(c). We observe that $p$ decreases (weights get sparser) until $M$ reaches 50--100.

For values of $M > 100$, the modes of covariation cannot be decomposed into thinner ones anymore, and $p$ saturates to the minimal value $\sim 2/27$ corresponding to a hidden unit linked to two visible units (Fig.~\ref{compositional_figures}(a)). Then, additional features are essentially duplicates of the previous ones. This can be seen from the distribution of maximum weight overlaps $O^{\max}_\mu $ shown in Fig.~\ref{compositional_figures}(a), see Appendix for definition. 

In addition, the number $L$ of simultaneously active hidden units per sequence grows, undergoing a continuous transition from a ferromagnetic-like regime ($L\sim 1$) to a compositional one ($L \gg 1$). Note crucially that $L$ does not scale linearly with $M$, see Fig.~\ref{compositional_figures}(a,b). If we account for duplicated hidden units, $L$ tends to saturate at about $12$, a number that arises from the data distribution rather than from $M$. In contrast, for an unregularized training with quadratic hidden unit potential, $L$ grows to much larger values ($L\sim 50$) as $M$ increases (Appendix, Fig.~\ref{compositional_figures_sup}). Lastly, Fig. \ref{compositional_figures}(c) shows that the theoretical scaling $L \sim \frac{1}{p}$ \cite{tubiana2017emergence} is qualitatively correct.

%
% comparison with other approaches
%
\section{Comparison with other Representation Learning Algorithms}\label{sec_comparison}

\begin{figure}
\includegraphics[scale=0.8]{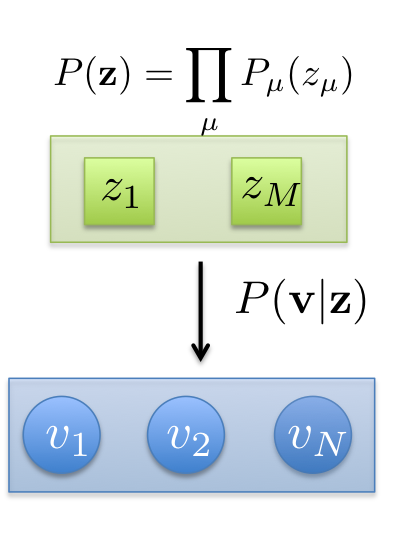}
\caption{Directed latent variable model. Latent variables are first drawn independently from one another, then a configuration is sampled from $P({\bf v} | {\bf z})$, see Section IV.A}
\label{directed_model}
\end{figure}

\subsection{Models and definitions}
We now compare the results obtained with standard representation learning approaches.  Thanks to popular packages such as Scikit-learn or Keras, most of these approaches are easy to implement in practice.

Principal Component Analysis (PCA) is arguably the most commonly used tool for finding the main modes of covariation of data. It is routinely applied in the context of protein sequence analysis for finding protein subfamilies, identifying specificity-determining positions \cite{casari1995method,rausell2010protein,de2013emerging}, and defining subsets of sites (called sectors) within proteins that control the various properties of the protein % defining subsets of sites (called sectors) within proteins that evolve independently from one another and control the various properties of the protein, such as stability, binding affinity and specificity
\cite{halabi2009protein,mclaughlin2012spatial}. A major drawback of PCA is that it assumes that the weights are orthogonal, which is in general not true and often results in extended modes that cannot be interpreted easily. Independent Component Analysis (ICA) \cite{bell1995information,hyvarinen2000independent} is another approach that aims at alleviating this issue by incorporating high-order moments statistics into the model; ICA was applied for identifying neural areas from fMRI data \cite{mckeown1998analysis}, and also used for protein sequence analysis \cite{rivoire2016evolution}. Another way to break the rotational invariance is to impose sparsity constraints on the weights or the representation via regularization. Sparse PCA \cite{zou2006sparse} is one such popular approach, and was considered both in neuroscience \cite{baden2016functional} and protein sequence analysis \cite{quadeer2018co}. We will also study sparse (in weights) single-layer noisy autoencoders, which can be seen as a nonlinear variant of sparse PCA. Sparse dictionaries \cite{olshausen1996emergence,mairal2009online} are also considered. Finally, we will consider Variational Autoencoders \cite{kingma2013auto} with a linear-nonlinear decoder, which can be seen both as regularized autoencoder and as a nonlinear generative PCA. VAE were recently considered for generative purpose in proteins \cite{novak,marks,greener2018design}, and their encoder defines a representation of the data.

All the above mentioned model belong to the same family, namely of the linear-nonlinear latent variable graphical models, see Fig.~\ref{directed_model}. In this generative model, latent factors $z_\mu$ are drawn from an independent distribution $P({\bf z}) = \prod_\mu P_\mu(z_\mu)$, and the data is obtained, as in RBM, by a linear transformation followed by an element-wise nonlinear stochastic or deterministic transformation, of the form $P({\bf v} | {\bf z}) = \prod_i P(v_i | I_i^v({\bf z}))$. Unlike in RBM, a single pass is sufficient to sample configurations rather than an extensive back-and-forth process. For a general choice of $P_\mu(z_\mu)$ and $P({\bf v}|{\bf z})$, the  inference of this model by maximum likelihood is intractable because the marginal $P({\bf v})$ does not have a closed form. The different models correspond to different hypothesis on $P_\mu(z_\mu)$, $P\left( v_i | I_i^v({\bf z}) \right)$, and learning principles that simplify the inference problem, see Table~\ref{table_directed_model}.

\begin{table*}[t]
\begin{tabular}{|r|r|r|r|r|r|r|}
\hline
Algorithm & PCA & ICA (Infomax) & Sparse PCA & Sparse noisy & Sparse Dictionaries & Variational\\
  &  &  & & Autoencoders & & Autoencoders \\
 \hline
$P_\mu(z_\mu)$ & Gaussian & Non-Gaussian & / & / & Sparse & Gaussian\\
$M$ & $\leq N$ & $=N$ & $\lessgtr N$ & $\lessgtr N$ & $\lessgtr N$  & $\lessgtr N$ \\
$P(v_i | I_i^v)$ & Deterministic & Deterministic & Deterministic & Deterministic & Deterministic & Stochastic\\
$P({\bf z} | {\bf v})$ & Deterministic & Deterministic & Deterministic & Stochastic & Deterministic & Stochastic\\
$\left\langle v_i \right\rangle(I_i^v)$ & Linear & Linear  & Linear & Softmax & Linear & Softmax\\
$w_{i\mu}(v)$ & Orthonormal & Normalized & Sparse & Sparse & Normalized & Sparse \\ 
 \hline
Learning & Max. Likelihood & Max. Likelihood & Min. Reconstruction & Min. Noisy & Min. Reconstruction & Variational \\
Method & $\Leftrightarrow$ Min. Reconstruction & & Error (MSE) & Reconstruction  & Error (MSE)  & Max. Likelihood\\
& Error (MSE) & &  & Error (CCE)& & \\

% & $\Leftrightarrow$ Min. Reconstruction error (MSE) & Max. Lik & Min. MSE/CCE & Variational Max. Lik.\\

\hline
\end{tabular}
\caption{Similarities and differences between various feature extraction approaches. Abbreviations: MSE = Mean Square Error. CCE: Categorical Cross-entropy.}
\label{table_directed_model}
\end{table*}

\subsection{Lattice Proteins: Features inferred}

\begin{figure*}
\vspace*{-0.8in}
\includegraphics[scale=1.0]{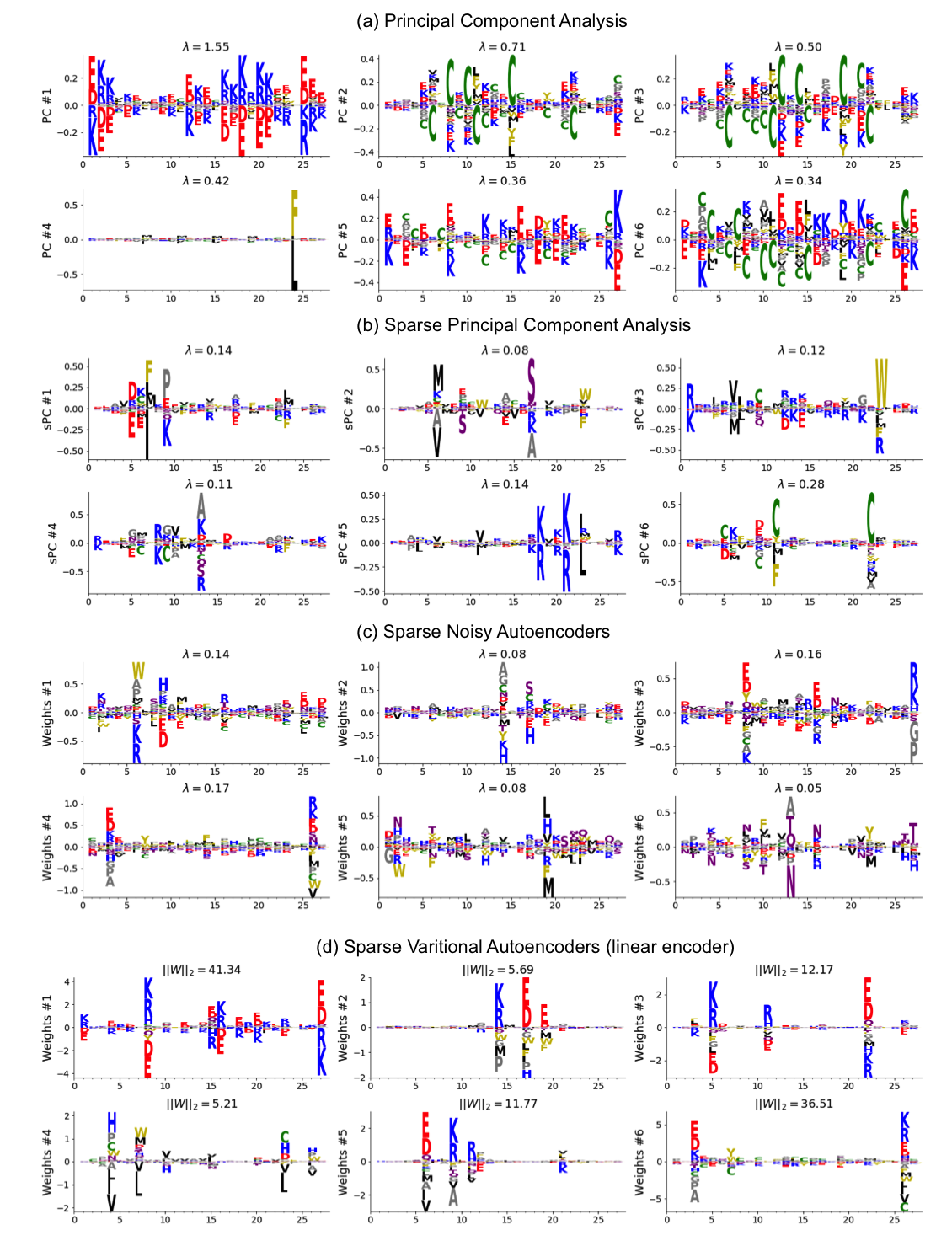}
\vspace*{-0.2in}
\caption{Six weights attached to latent factors for (a) Principal Component Analysis, (b) Sparse Principal Component Analysis, (c) Sparse Noisy Autoencoders and (d) Sparse variational autoencoders with linear encoding. Weights are selected as follows: (a) the first 6 principal components. (b),(c),(d) Selected non-zero weights with weight sparsity $p_\mu$ close to the median value. Values above the components indicate the feature importance, measured either by the latent factor variance $\lambda$ (when the weights are normalized) or the weight norm $\|W\|_2$ (when the variances are normalized).}
\label{figure_other_features}
\end{figure*}

We use each approach to infer features from the MSA Lattice Proteins. For each model, we use the same number $M=100$ of latent factors. For PCA, ICA and Sparse Dictionaries, we used one-hot-encoding for each site  (\textit{i.e.} converted into a 20-dimensional binary vector), and the standard implementation from Scikit-learn with default parameters. For sparse dictionaries, we adjusted the penalty such that the number $L$  of simultaneously active latent features matches the one found in RBM $\sim 10$. Sparse PCA and sparse autoencoders are special cases of autoencoders, respectively with a mean square reconstruction error and categorical cross-entropy reconstruction error, and were implemented in Keras. In both cases we used the same weights for encoding and decoding. For the noisy variant, we replace 20\% of the amino acids with a random one before attempting to reconstruct the original sequence. For Variational Autoencoders, we used the same parametric form of $P({\bf v} | {\bf z})$ as for RBM, \textit{i.e.} a linear transformation followed by a categorical distribution. The posterior probability $P({\bf z} | {\bf v} )$ is approximated by a Gaussian encoder $\mathcal{N}\left( \mu({\bf v}), \sigma^2({\bf v}) \right)$ where $\mu({\bf v})$ and $\log \sigma^2({\bf v})$ are computed from ${\bf v}$ either by linear transformation or by a single hidden layer neural network; the parameters of both $P({\bf v} | {\bf z})$ and the encoder are learnt by maximizing the variational lower bound on the likelihood of the model. The sparse regularization is enforced on the 'decoding' weights of $P({\bf v} | {\bf z})$ only. For all models with a sparse weight regularization, we selected the regularization strength so as to obtain similar median sparsity values $p$ as with the RBM shown in Fig.~\ref{RBM_Features_LP}. We show for each method six selected features in Fig.~\ref{figure_other_features}  (PCA, sPCA, sNAE, sVAE with linear encoder), and in Appendix Fig.~\ref{figure_other_features_sup} (ICA, Sparse Dictionaries, sVAE with nonlinear encoder). For PCA, ICA and Sparse Dictionaries, we show the most important features in terms of explained variance. For sPCA, sNAE and sVAE, we show features with sparsity $p_\mu$ close to the median sparsity.

Most of the high importance features inferred by PCA, ICA are completely delocalised and encode the main collective modes of the data, similarly to unregularized RBM. Clearly, there is no simple way to relate these features to the underlying interactions of the system. Even if sparsity could help, the main issue is the enforced constraint of decorrelation/independance, because it impedes the model from inferring smaller coevolutionary modes such as pairs of amino acids in contact. Indeed, Lattice Proteins exhibits a hierarchy of correlations, with sites that are tightly correlated, and also weakly correlated to others. For instance, hidden units 4 and 5 of Fig.~\ref{RBM_Features_LP} are strongly but not completely correlated, with a small fraction of sequences having $I_4>0$ and $I_5<0$ and conversely (Appendix Fig.~\ref{figure_latent_representation}(a)). Therefore, neither PCA nor ICA can resolve the modes; instead, the first principal component is roughly the superposition of both modes (Fig.~\ref{fig_LP}(e)). Finally, both PCA and ICA also infer a feature that focuses only on site 24 (PC4 and IC4). As seen from the conservation profile (Fig.~\ref{fig_LP}(d)), this site is strongly  conserved, with two possible amino acids. Since mutations are uncorrelated from the remaining of the sequence and the associated variance of the mode is fairly high, both PCA and ICA encode this site. In contrast, we never find a hidden unit focusing only on a single site with RBM, because its effect on $P({\bf v})$ is equivalent to a field term. Similar features are found with Sparse Dictionaries as well, see Appendix.

As expected, all the models with sparse weights penalties (sPCA, sNAE and sVAE) can infer localized features, provided the regularization is adjusted accordingly. However, unlike in RBM, a significant fraction of these features does not focus on sites in contact. For instance, in sparse PCA the features 2 and 5 focus respectively on pairs 6-17 and 17-21, and neither of them are in contact. To be more systematic, we identify for each method the features focusing on two and three sites (via the criterion $\sum_v |w_{i\mu}(v)| > 0.5 \max_i  \sum_v |w_{i\mu}(v)|$), count them and compare them to the original structure. Results are shown in Table~\ref{table2}. For RBM, 49 hidden units focus on two sites, of which 47 are in contact, and 18 focus on three sites, of which 16 are in contact (e.g. like 8-16-27). For the other methods, both the number of features and the true positive rate are significantly lower. In Sparse PCA, only 15/25 pairs and 3/24 triplet features are legitimate contacts or  triplets in the structure.

\begin{table}[t]
\begin{tabular}{|r|r|r|}
\hline
Sparse Model & Pair features & Triplet features \\
\hline
RBM & 47/49 & 16/18 \\
sPCA & 15/25 & 3/24 \\
Noisy sAE & 20/29 & 7/19 \\
sVAE (linear encoder) & 8/8 & 5/6 \\
sVAE (nonlinear encoder) & 6/6 & 2/4 \\ 
\hline
\end{tabular}
\caption{Number of sparse features extracted by the various approaches that are localized on the structure $S_A$.}
\label{table2}
\end{table}

The main reason for this failure is that in sparse PCA (as in PCA), emphasis is put on the complete reconstruction of the sequence from the representation because the mapping is assumed to be deterministic rather than stochastic. The sequence must be compressed entirely in the latent factors, of smaller size $M=100 < 27\times 20=540$, and this is achieved by ‘grouping’ sites in a fashion that may respect some correlations, but not necessarily the underlying interactions. Therefore, even changing the reconstruction error to cross-entropy to properly take into account the categorical nature of the data does not significantly improve the results. However, we found that corrupting the sequence with noise before attempting to reconstruct it (i.e. introducing a stochastic mapping) indeed slightly improves the performance, though not to the same level as RBM for quantitative modeling.

The simplifying assumption that $P({\bf v} | {\bf z} )$ is deterministic can be lifted by adopting a variational approach for learning the model; in the case of Gaussian distribution for $z_\mu$ we obtain the Variational Autoencoder model. The features obtained are quite similar to the ones obtained with RBM, featuring contacts, triplets (with very few false positives) and some extended modes. Owing to this stochastic mapping, the representation need not encode individual site variability, and focuses instead on the collective behavior of the system.

The major inconvenient of VAE is that we find that only a small number of latent factors ($\sim 20$) are effectively connected to the visible layer. Increasing the number of latent factors or changing the training algorithm (SGD vs ADAM, Batch Normalization) did not improve this number. This is likely because the independent Gaussian assumption is not compatible with the linear-nonlinear decoder and sparse weights assumption. Indeed, the posterior distribution of the latent factors (from $P({\bf z} | {\bf v})$) show significant correlations and deviations from Gaussianity (Appendix Fig.~\ref{figure_latent_representation}(b,c)); The KL regularization term of VAE training therefore encourages them from being disconnected. Another consequence of this deviation from i.i.d. Gaussian is that sequences generated with the VAE have significantly lower fitness and diversity than with RBM (Appendix Figs.~\ref{generated_LinearVAE},\ref{generated_nonLinearVAE}). Though deep mappings between the representation and the sequence (as in \cite{marks}), as well as more elaborate priors (as in \cite{mescheder2017adversarial}) can be considered for quantitative modeling , the resulting models are significantly less interpretable. In contrast, undirected graphical models such as RBM have a more flexible latent variable distribution, and are therefore more expressive for a fixed linear-nonlinear architecture. 

In summary, the key ingredients allowing to learn RBM-like representations are i) Sparse weights ii) Stochastic mapping between configurations and representations and iii) Flexible latent variable distributions.

\subsection{Lattice Proteins: Robustness of features}
\begin{figure*}
\includegraphics[scale=1.0]{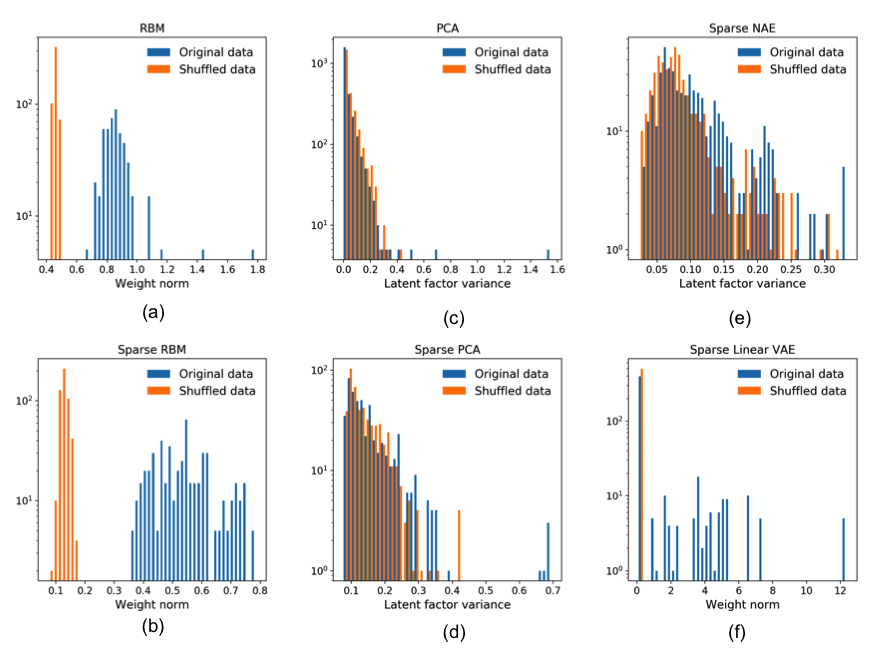}
\caption{Comparison of feature importance distribution for real (blue) and shuffled (orange) data, for (a) unregularized RBM, (b) regularized RBM, (c) PCA, (d) Sparse PCA, (e) Sparse Noisy Autoencoders, (f) Sparse Variational Autoencoders (with linear encoding)}
\label{figure_statistical_efficiency}
\end{figure*}

\begin{figure*}
\hspace*{-0.6in}
\includegraphics[scale=1.2]{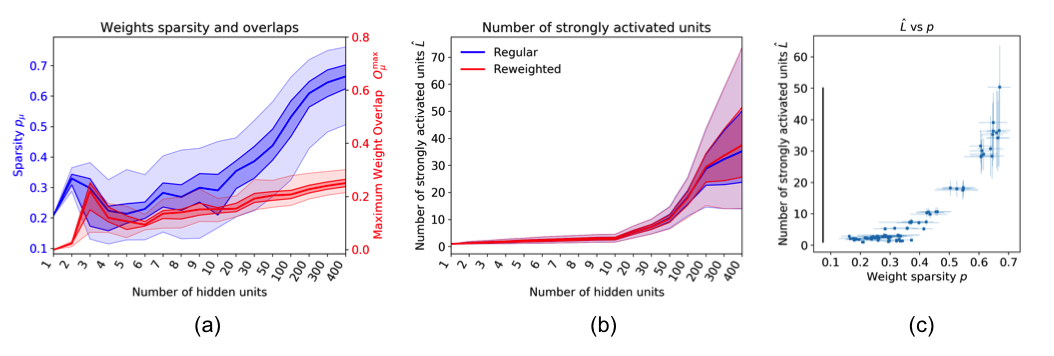}
\caption{Evolution of the weight sparsity $p$ as a function of the number of hidden units, for quadratic potential and no regularization. Unlike in the compositional phase, the number of activated hidden units scales as $M$}
\label{compositional_figures_sup}
\end{figure*}

Somewhat surprisingly, we also find that allowing a stochastic mapping between representation and configurations results in features that are significantly more robust with respect to finite sampling. For all the models and parameters used above, we repeat the training five times with either the original data set and varying seed or the shuffled data set, in which each column is shuffled independently from the others so as to preserve the first order moments and to suppress correlations. For a sample of very large size, the models trained on the shuffled data should have latent factors that are either disconnected or localized on a single site, with small feature importance. We show in Fig.~\ref{figure_statistical_efficiency} and Appendix Fig.~\ref{figure_statistical_efficiency_sup} the distribution of feature importance for the original and shuffled data, for each method. For PCA, ICA, sparse PCA and sparse autoencoders, the weights are normalized and feature importance is determined by the variance of the corresponding latent factor. For RBM and VAE, the variance of the latent factors is normalized to 1 and the importance is determined by the weight amplitude.  
In PCA and ICA, we find that only a handful of features emerge from the bulk; sparse regularization slightly improves the number but not by much. In contrast, there is a clean scale separation between feature importance for regular and shuffled data, for both regularized and unregularized RBM and VAE. This notably explains why only few principal components can be used for protein sequence modelling, whereas many features can be extracted with RBM. The number of modules that can be distinguished within protein sequences is therefore grossly underestimated by Principal Component Analysis.

\section{Conclusion}
Understanding the collective behavior of a physical system from observations requires to learn a phenomenology of data, as well as to infer the main interactions that drive its collective behavior. Representation learning approaches, such as as Principal Component Analysis or Independent Component Analysis have been frequently considered for this purpose. Comparatively, Restricted Boltzmann Machines, a Machine Learning tool originally popularized for unsupervised pretraining of Deep Neural Networks have received little attention, notably due to a limited of understanding of the properties of the learnt representations.

In this work, we have studied in detail the performances of RBM on a notoriously difficult problem, that is, the prediction of the properties of proteins from their sequence (the so-called genotype-to-phenotype relationship). We have shown that, provided that appropriate non-quadratic potentials, as well as sparsity regularization over the weights are used, RBM can learn compositional representations of data, in which the hidden units code for constitutive parts of the data configurations. Each constitutive part, due to sparsity, focuses on a small region of the data configurations (subset of visible sites), and is therefore easier to interpret that extended features. Constitutive parts may, in turn, be stochastically recombined to generate new data with good properties (here, probability of folding into the desired 3D structure).  Regularized RBM therefore offer an appealing compromise between model interpretability and generative performance.

We stress that the behavior of RBM described above is in full agreement with theoretical studies of RBM with weights extracted from  random statistical ensembles, controlled by a few parameters, including the sparsity, the aspect ratio of the machine, the thresholds of rectified linear units \cite{tubiana2017emergence}. It is however a non trivial result, due to the very nature of the lattice-protein sequence distribution, that forcing RBM trained on such data to operate in the compositional regime can be done at essentially no cost in log-likelihood (see Fig.~\ref{sparsity_performance}(b)). This property also holds for real proteins, as shown in \cite{tubiana2018learning}.

In addition, RBM enjoys some useful properties with respect to the other representation learning approaches studied in the present work. First, RBM representations focus on the underlying interactions between components rather than on all the variability of the data, taken into account by site-dependent potentials acting on the visible unit. The inferred weights are therefore fully informative on the key interactions within the system. Secondly, the distribution of latent factors (Fig.~\ref{directed_model}) is not imposed in RBM, contrary to VAE where it is arbitrarily supposed to be Gaussian. The inference of the hidden-unit potential parameters (thresholds $\theta_\pm$ and curvatures $\gamma_\pm$ for dReLU units) confers a lot of adaptibility to RBM to fit as closely as possible the data distribution. 

Altogether, beyond the protein sequence analysis application presented here, our work suggests that RBM shed a different light on data and could find usage to model other data in an accurate and interpretable way. It would be very interesting to extend this approach to deeper architectures, and see how the nature of representations vary from layer to layer. Another possible extension regards sampling. RBM, when driven in the compositional phase, empirically show efficient mixing properties. Characterizing how fast the data distribution is dynamically sampled would be very interesting, with potential payoff for training where benefiting from efficient sampling is crucial.

\subsection*{Acknowledgments}
We acknowledge Cl\'ement Roussel for helpful comments. This work was partly funded by the ANR project RBMPro CE30-0021-01. J.T. acknowledges funding from the Safra Center for Bioinformatics, Tel Aviv University.

\bibliography{bibliography.bib}

\beginsupplement
\section*{Annex: Proxies for weight sparsity, number of strongly activated hidden units}\label{app_p}

Since neither the hidden layer activity nor the weights are exactly zero, proxies are required for evaluating them. In order to avoid the use of arbitrary thresholds which may not be adapted to every case, we use Participation Ratios.

The Participation Ratio $(PR_e)$ of a vector ${\bf x}=\{x_i \}$ is:

\begin{equation}
PR_e(\bf x) =  \frac{(\sum_{i} |x_i|^e)^2}{\sum_{i} |x_i|^{2e} }
\end{equation}

\noindent If $\bf x$ has $K$ nonzero and equal (in modulus) components PR is equal to $K$ for any $a$. In practice we use the values $a=2$ and 3: the higher $a$ is, the more small components are discounted against strong components in $\bf x$. Note also that it is invariant to rescaling of ${\bf x}$.

PR can be used to estimate the weight sparsity for a given hidden unit, and averaged to get a single value for a RBM.

\begin{equation}\label{sparsity}
\begin{split}
w^S_{i\mu} \equiv \sqrt{\sum_a w_{i\mu}(a)^2} \\
p_\mu = \frac{1}{N} PR_2( \bf{w^S_\mu}) \\
p = \frac{1}{M} \sum_\mu p_\mu
\end{split}
\end{equation}

Similarly, the number of strongly activated hidden units for a given visible layer configuration can be computed with a participation ratio. For non-negative hidden units such as with the ReLU potential, it is obtained via:

\begin{equation}
\begin{split}
h_\mu = \left\langle h_\mu | I_\mu({\bf v}) \right\rangle \\
L = PR_3({\bf h_\mu} )
\end{split}
\end{equation}

For dReLU hidden units, which can take both positive and negative values and may have e.g. bimodal activity, their most frequent activity can be non-zero. We therefore subtract it before computing the participation ratio:

\begin{equation}
\begin{split}
h_\mu = \left\langle h_\mu | I_\mu({\bf v}) \right\rangle \\
h_\mu^0 = \arg \max P_{\text{data}}(h_\mu) \\
L = PR_3({\bf |h_\mu - h_\mu^0|} )
\end{split}
\end{equation}

To measure the overlap between hidden units $\mu,\mu'$, we introduce the quantities
\begin{equation}
O_{\mu \nu} = \frac{\sum_{i,v} w_{i\mu}(v) w_{i\nu}(v) }{\sqrt{\left(\sum_{i,v} w_{i\mu}(v)^2\right) \left(\sum_{i,v} w_{i\nu}(v)^2 \right)}} 
\ ,
\end{equation}
which takes values in the $[-1,1]$ range. We also define $O^{\max}_\mu = \max_{\mu'\neq \mu} |O_{\mu,\mu'}|$.

To account for strongly overlapping hidden units, one can also compute the following weighted participation ratio:
\begin{equation}
PR_e(\bf h, {\bf w}) =  \frac{(\sum_{\mu} w_\mu |h_\mu|^e)^2}{\sum_{\mu} w_\mu |h_\mu|^{2e} }
\end{equation}
where $w_\mu$ is chosen as the inverse of the number of neighboring hidden units, defined according to the criterion $|O_{\mu\nu}|>0.9$. This way, two hidden units with identical weights contribute to the participation ratio as much as a single isolated one.

\section*{Annex: Supplementary figures}

\begin{figure*}
\vspace*{-0.2in}
\includegraphics[scale=1.0]{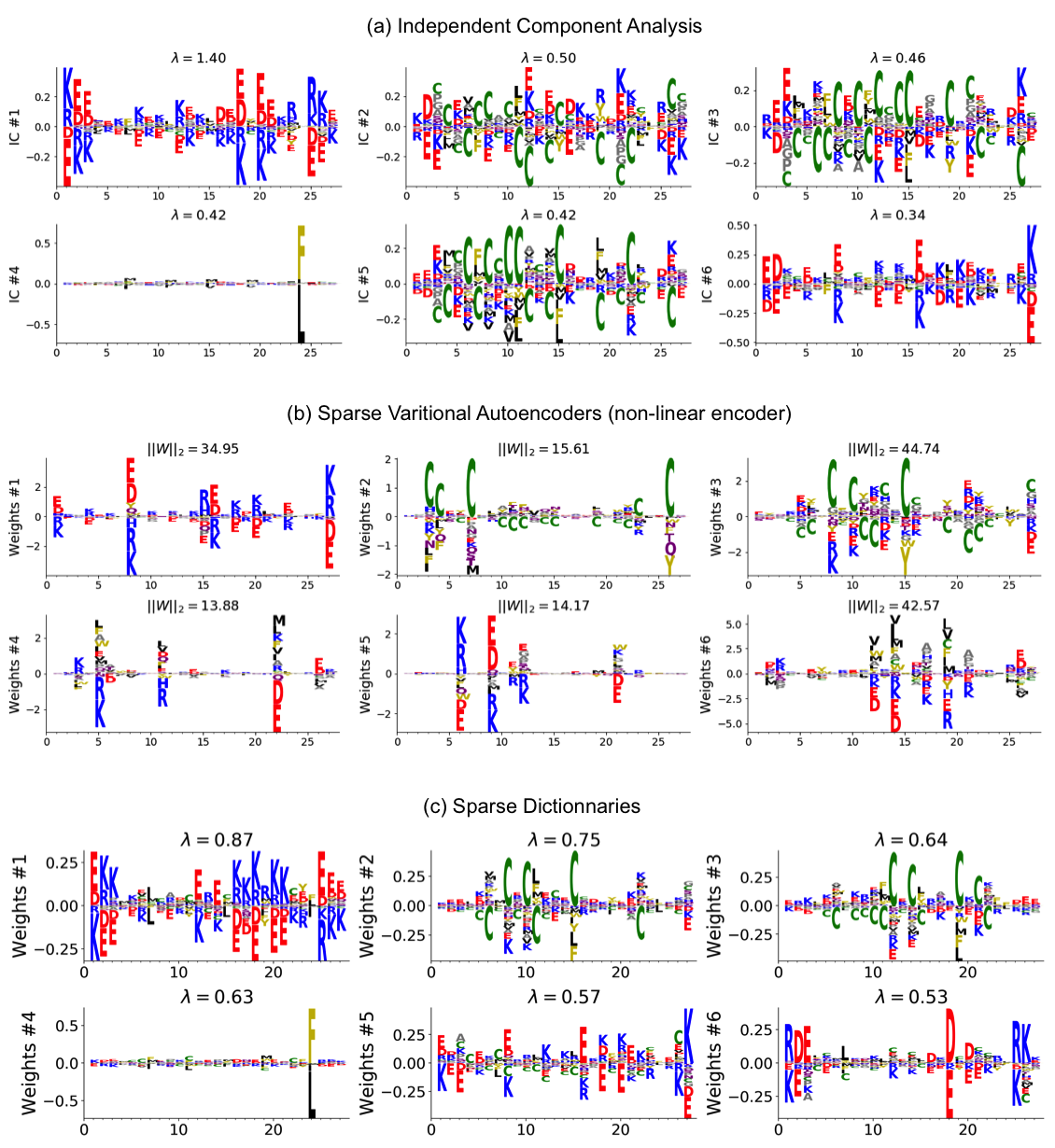}
\vspace*{-0.2in}
\caption{Six weights attached to latent factors for (a) Independent Component Analysis, (b) Sparse variational autoencoders with nonlinear encoding (c) Sparse dictionaries. For (a),(c) the six latent factor with largest variance are shown. For (b), we selected six latent factors whose corresponding weights have weight sparsity $p_\mu$ close to the median value.}
\label{figure_other_features_sup}
\end{figure*}

\begin{figure*}
\includegraphics[scale=1.0]{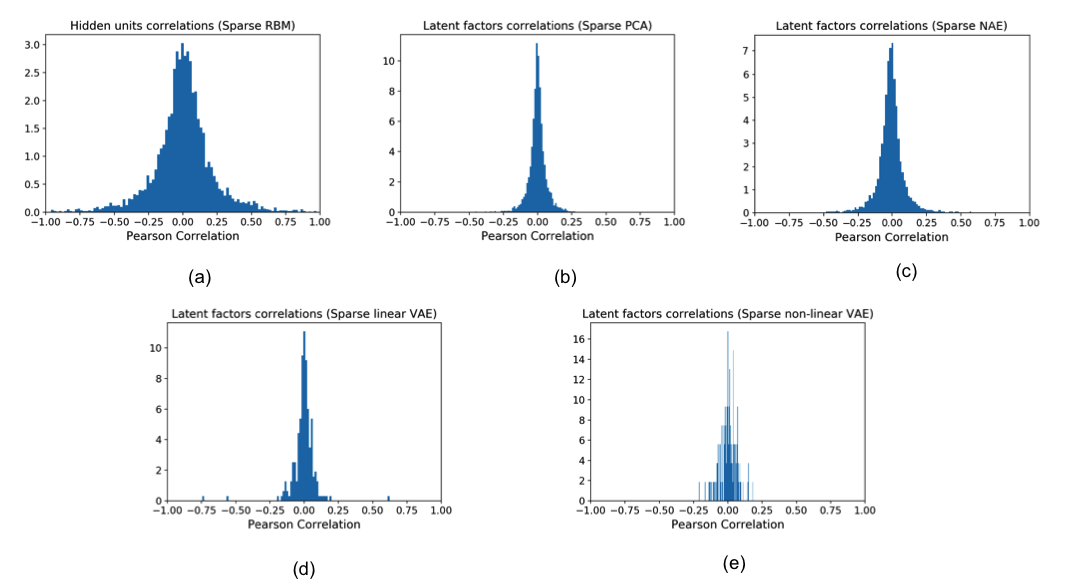}
\caption{Distribution of Pearson correlations within the representation for the models shown in Fig.~\ref{figure_other_features}, \ref{figure_other_features_sup}. (a) RBM (regularized, $\lambda_1^2=0.025$), (b) Sparse PCA (c) Sparse noisy Autoencoders (d) Sparse VAE with linear encoder (e) Sparse VAE with nonlinear encoder. For models (b)-(e), the sparse weight regularization is selected to yield same median weight sparsity as for the regularized RBM.}
\label{figure_correlations}
\end{figure*}

\begin{figure*}
\includegraphics[scale=1.0]{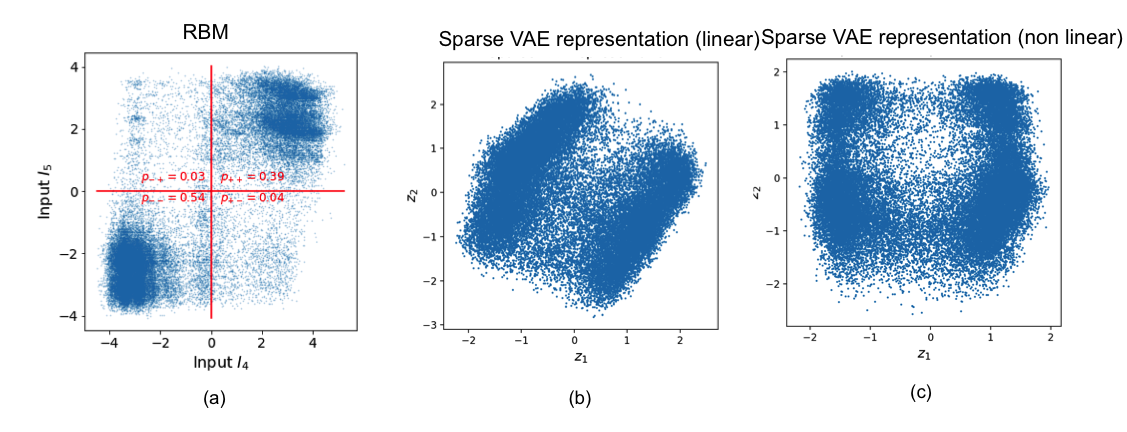}
\caption{Example of latent factor data distributions, i.e. obtained by sampling from $P( {\bf h}/{\bf z} | {\bf v})$ for each sequence ${\bf v}$ in the data. (a) Scatter plot of $h_4$ vs $h_5$ for the RBM shown in Fig.~\ref{RBM_Generation_LP}. Note the strong but imperfect correlation and the four distinct clusters. (b),(c) Scatter plot of $z_1$ vs $z_2$, the two most important features, for the Sparse VAE with linear (b) or nonlinear encoding (c). Note the strong deviations from the i.i.d. gaussian distribution.}
\label{figure_latent_representation}
\end{figure*}

\begin{figure*}
\includegraphics[scale=1.0]{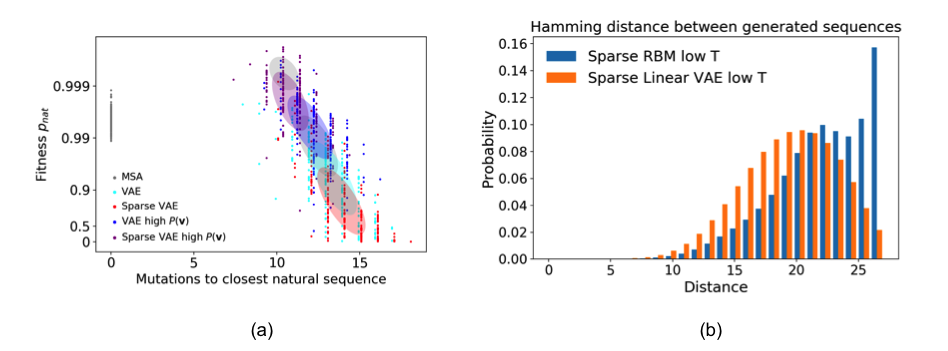}
\caption{{\bf Sequence generation with linear VAE} {\bf (a)} Scatter plot of the number of mutations to the closest natural sequence vs probability to fold into $S_A$, for natural  and artificial LP sequences. Sequences are generated from unregularized and regularized VAE, by sampling random gaussian iid ${\bf z}$, then ${\bf v}$ are obtained either by sampling from $P({\bf v} | {\bf z})$ or as $\arg\max_{\bf v} P({\bf v} | {\bf z})$ (high probability). Same colors and scale as Fig.~\ref{RBM_Generation_LP}. RBM sequences (regular and high $P({\bf v})$ are shown in gray. {\bf (b)} Distribution of Hamming distance between generated sequences, showing the diversity of the generated sequences}
\label{generated_LinearVAE}
\end{figure*}

\begin{figure*}
\includegraphics[scale=1.0]{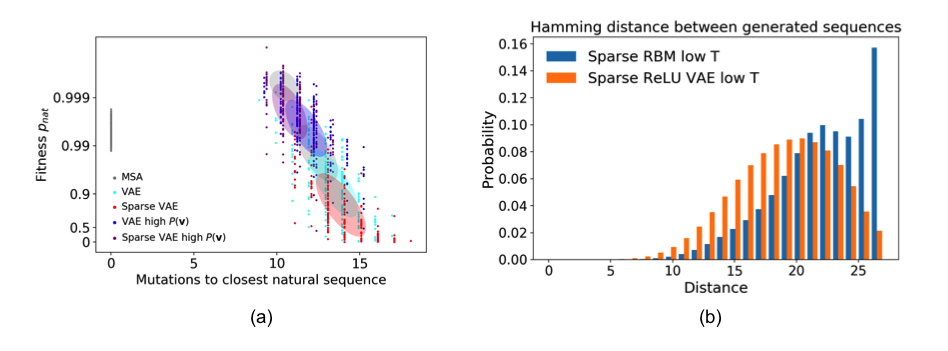}
\caption{Same as Supplementary Fig.~\ref{generated_LinearVAE} for VAE with  encoding.}
\label{generated_nonLinearVAE}
\end{figure*}

\begin{figure*}
\includegraphics[scale=1.0]{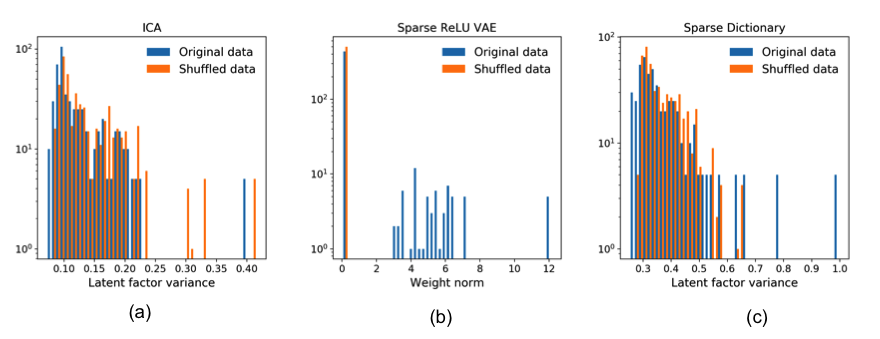}
\caption{Comparison of feature importance distribution for real (blue) and shuffled (orange) data, for (a) ICA, (b) Sparse Variational Autoencoders (with nonlinear encoding)}
\label{figure_statistical_efficiency_sup}
\end{figure*}

\end{document}